\def\doctype{0}
\documentclass[letterpaper]{article} % DO NOT CHANGE THIS
\usepackage{aaai24_preprint}  % DO NOT CHANGE THIS
\usepackage{times}  % DO NOT CHANGE THIS
\usepackage{helvet}  % DO NOT CHANGE THIS
\usepackage{courier}  % DO NOT CHANGE THIS
\usepackage[hyphens]{url}  % DO NOT CHANGE THIS
\usepackage{graphicx} % DO NOT CHANGE THIS
\usepackage{natbib}  % DO NOT CHANGE THIS AND DO NOT ADD ANY OPTIONS TO IT
\usepackage{caption} % DO NOT CHANGE THIS AND DO NOT ADD ANY OPTIONS TO IT
\usepackage{algorithmic}
\usepackage{graphicx}
\usepackage{xcolor}
\usepackage{amssymb}
\usepackage{graphicx}
\usepackage{dsfont}
\usepackage{subcaption}
\usepackage[ruled,vlined]{algorithm2e}
\usepackage[page]{appendix}
\usepackage[utf8]{inputenc}
\usepackage{amsmath}
\usepackage{caption}
\usepackage{resizegather}
\usepackage{multirow}
\newcommand{\risk}[1]{\varrho^{#1}}

\newcommand{\radius}{boundary\xspace}

\newtheorem{theorem}{Theorem}[section]

\newtheorem{remark}{Remark}[section]
\newtheorem{definition}{Definition}[section]

\newtheorem{lemma}{\textbf{Lemma}}
\newcommand{\E}{\mathbb{E}}

\DeclareMathOperator*{\argmin}{arg\,min}
\title{Safeguarded Progress in Reinforcement Learning:\\ Safe Bayesian Exploration for Control Policy Synthesis}
\author{
    Rohan Mitta\textsuperscript{\rm 1},
    Hosein Hasanbeig\textsuperscript{\rm 2}$^*$,
    Jun Wang\textsuperscript{\rm 3},\\
    Daniel Kroening\textsuperscript{\rm 4}\thanks{The work in this paper was done at the University of Oxford.},
    Yiannis Kantaros\textsuperscript{\rm 3},
    Alessandro Abate\textsuperscript{\rm 1}$^\dag$
}
\affiliations{
    \textsuperscript{\rm 1}University of Oxford,
    \textsuperscript{\rm 2}Microsoft Research,
    \textsuperscript{\rm 3}Washington University,
    \textsuperscript{\rm 4}Amazon\\
    {$^\dag$}alessandro.abate@cs.ox.ac.uk
}

\usepackage{bibentry}
\begin{document}

\maketitle

\begin{abstract}
This paper addresses the problem of maintaining safety during training in Reinforcement Learning (RL), such that the safety constraint violations are bounded at any point during learning. 
In a variety of RL applications the safety of the agent is particularly important, e.g. autonomous platforms or robots that work in proximity of humans. 
%Thus, researchers are paying increasing attention not only to maximise the long-term task-driven reward, but also to damage avoidance. 
As enforcing safety during training might severely limit the agent's exploration, we propose here a new architecture that handles the trade-off between efficient progress and safety during exploration. 
As the exploration progresses, we update via Bayesian inference Dirichlet-Categorical models of the transition probabilities of the Markov decision process that describes the environment dynamics. This paper proposes a way to approximate moments of belief about the risk associated to the action selection policy.   
We construct those approximations, and prove the convergence results. 
We propose a novel method for leveraging the expectation approximations to derive an approximate bound on the confidence that the risk is below a certain level. 
This approach can be easily interleaved with RL and we present experimental results to showcase the performance of the overall architecture.
\end{abstract}

\section{Introduction}

\if\doctype 2
Reinforcement Learning (RL) is a machine learning paradigm in which an agent interacts with a (generally unknown) environment to learn to achieve maximal long-term reward from that environment. The environment is typically modelled as a Markov Decision Process (MDP) with state/action-based rewards. The agent explores the environment by sequentially selecting actions that are fed into the MDP to generate next states and corresponding scalar rewards for the agent. Thus, as the agent explores the environment, it can incrementally learn a policy - namely, a mechanism for action selection - that maximises the expected reward from the environment~\citep{sutton_bach_barto_2018,puterman}. 
\fi

Traditionally, RL is principally concerned with the policy that the agent generates by the end of the learning process. In other words, the quality of agent's policy \textit{during} learning is overlooked at the benefit of learning how to behave optimally, eventually. Accordingly, many standard RL methods rely on the assumption that the agent selects each available action at every state infinitely often during exploration~\citep{sutton_bach_barto_2018,puterman}. A~related technical assumption that is often made is that the MDP is \textit{ergodic}, meaning that every state is reachable from every other state under proper action selection~\citep{moldovan_safe_2012}. These assumptions might be reasonable, e.g., in virtual environments where restarting is always an option. However, in safety-critical scenarios these assumptions might be unreasonable, as we may explicitly require the agent to never visit certain unsafe states. Indeed, in a variety of RL applications the safety of the agent is particularly important, e.g., when using expensive autonomous platforms or robots that work in the proximity of humans. Thus, researchers are recently paying increasing attention not only to maximising a long-term task-driven reward, but also to enforcing safety during training. 

\textbf{Related Work } 
The general problem of \emph{Safe} RL has been an active area of research in which numerous approaches and definitions of safety have been proposed~\citep{Brunka_Greeff_Hall,garcia_comprehensive_nodate,risk4}. \citet{moldovan_safe_2012} define safety in terms of ``actions availability'', namely ensuring that an agent is always able to return to its current state after moving away from it.   \citet{chow_risk-constrained_nodate} pursue safety by minimising a cost associated with worst-case scenarios, when cost is associated with a lack of safety. Similarly, \citet{miryoosefi2019reinforcement} define the safety constraint in terms of the expected sum of a vector of measurements to be in a target set.
Other approaches~\citep{lcrl2018,hasanbeig2019logicallyconstrained,lcnfq,lcddpg,hasanbeig2022certified,li2019temporal,deepsynth,cai2021modular,hasanbeig2023certified,hasanbeig2023symbolic} define safety by the satisfaction of temporal logical formulae of the learnt policy, but do not provide safety \textit{while} training such a policy. Many existing approaches have been concerned with providing guarantees on the safety of the learned policy often under the assumption that a backup policy is available~
\citep{de2019foundations,de2020restraining,coraluppi_marcus_1999,perkins2002lyapunov,geibel_wysotzki_2005,mannucci2017safe,lyap,mao2019towards}.
These methods are applicable to systems if they can be trained on accurate simulations, but for many other real-world systems we instead require safety \textit{during} training. 

There has also been much research done into the development of approaches to maintaining safety during training. For instance, \citep{alshiekh2017safe,jansen2019safe,bounded_shield} leverage the concept of a \emph{shield} that stops the agent from choosing any unsafe actions. The shield assumes the agent observes the entire MDP (and any opponents) to construct
a safety (game) model, which will be unavailable for many partially-known MDP tasks. The approach by \citet{Garcia_2012} assumes a predefined safe baseline policy that is most likely sub-optimal, and attempts to slowly improve it with a slightly noisy action-selection policy, while defaulting to the baseline policy whenever a measure of safety is exceeded. However, this measure of safety assumes that nearby states have similar safety levels, which may not always be the case. Another common approach is to use expert demonstrations to attempt to learn how to behave safely~\citep{abbeel_coates_ng_2010}, or even to include an option to default to an expert when the risk is too high~\citep{ALA12-Torrey}. Obviously, such approaches rely heavily on the presence and help of an expert, which cannot always be counted upon. Other approaches to this problem \citep{entropy,barrier_1,knownD2} are either computationally expensive or
require explicit, strong assumptions about the model of agent-environment interactions. Crucially, maintaining safety in RL by efficiently leveraging available data is an open problem~\citep{taylor2021towards}.

%\vspace{-5mm}\subsubsection*{Contributions - } 
%\edit{[if space is needed, we could make the rest of this section shorter, crisper:]} 
\textbf{Contributions } 
%In this work, we incorporate a number of elements of the approach to Safe RL proposed in~\citep{hasanbeig_cautious_2020}: 
%Parts of the problem setup introduced in the paper, and in particular some areas where we felt the paper went astray, were used as inspiration for the method proposed in this dissertation. 
Extending upon~\cite{hasanbeig_cautious_2020,cautious_bayesian_rl}, we tackle the problem of synthesising a policy via RL that 
optimises a discounted reward, 
%satisfies, with maximal probability, a Linear Temporal Logic (LTL) that specifies a safety requirement, 
while not violating a safety requirement \textit{during} learning. 
%\textcolor{red}{[this is the case in thesis but we haven't talked about LTL in this paper]} 
%or part of the global property.  
%We consider in particular their approach to ensuring satisfaction of a safety requirement during learning.  
%namely to avoid certain states classified as \textit{unsafe}. 
%In fact if one allows for a modification of the MDP, then we can always translate the problem of satisfying such a safety requirement into a problem of avoiding certain unsafe states. 
% In order to remain safe during learning,  \citep{hasanbeig_cautious_2020} 
This paper puts forward a \emph{cautious RL formalism} that assumes the agent (1) has limited observability over states and (2) infers a Dirichlet-Categorical model of the MDP dynamics. 
We incorporate higher-order information from the Dirichlet distributions, in particular we compute approximations of the (co)variances of the risk terms. %$p^{ij}_a$: 
This allows the agent to reason about the contribution of epistemic uncertainty to the risk level, and therefore to make better informed decisions about how to stay safe during learning. We show convergence results for these approximations, and propose a novel method to derive an approximate bound on the confidence that the risk is below a certain level. 
%The MDP model and its approximated distributions, and the associated measures of risk, are updated dynamically as new data is gathered during exploration. 
The new method adds a functionality to the agent that prevents it from taking critically risky actions, and instead leads the agent to take safer actions whenever possible, but otherwise leaves the agent to explore.  
The proposed method is versatile given that it can be added on to any general RL training scheme, in order to maintain safety during learning. 
%Taking inspiration from how humans learn in risky situations, we will assume the agent has a prior belief about the transition probabilities of the MDP and that the agent can observe which states in a neighbourhood of its current state are unsafe. Based on its beliefs,
%which are updated as it gathers new experience, 
%the agent can reason about the probability of entering such states, and decide whether or not an action is too risky to take.
Instructions on how to execute all the case studies in this paper are provided on the following GitHub page:
\begin{center}
    \url{https://github.com/keeplearning-robot/riskawarerl}
\end{center}

\section{Background}\label{sec:QL}

\subsection{Problem Setup }
\begin{definition}
A finite MDP with rewards~\citep{sutton_bach_barto_2018} is a tuple $M = \left<S, A, s_0, P, R\right>$ where $S = \{s^1,s^2,s^3,...,s^N\}$ is a finite set of states, $A$ is a finite set of actions, without loss of generality $s_0$ is an initial state, $P(s'|s,a)$ is the probability of transitioning from state $s$ to state $s'$ after taking action $a$, and $R(s,a)$ is a real-valued random variable which represents the reward obtained after taking action $a$ in state $s$. A realisation of this random variable (namely a sample, obtained for instance during exploration) will be denoted by $r(s,a)$. 
\end{definition}

An agent is placed at $s_0 \in S$ at time step $t = 0$. At every time step $t \in \mathbb{N}_0$, the agent selects an action $a_t \in A$, and the environment responds by moving the agent to some new state $s_{t+1}$ according to the transition probability distribution, i.e., $s_{t+1} \sim P(\cdot | s_t,a_t)$. The environment also assigns the agent a reward $r(s_t,a_t)$. The objective of the agent is to learn how to maximise the long term reward. In the following we explain these notions more formally. 

\begin{definition}
A policy $\pi$ assigns a distribution over $A$ at each state: $\pi(a|s)$ is the probability of selecting action $a$ in state $s$. Given a policy $\pi$, we can then define a state-value function 
$$
v_\pi(s) = \E^\pi \left[ \left. \sum_{t=0}^\infty \gamma^t r(s_t,a_t) \right| s_0 = s \right],
$$ 
where $\E^\pi[\cdot]$ is the expected value given that actions are selected from $\pi$, and $0<\gamma\leq1$ is a discount factor. 
\end{definition}
Specifically, this means that the sequence $s_0,a_0,s_1,a_1,...$ is such that $a_n\sim\pi(\cdot| s_n)$ and $s_{n+1}\sim P(\cdot|s_n,a_n)$. The discount factor $\gamma$ is a pre-determined hyper-parameter that causes immediate rewards to be worth more than rewards in the future, as well as ensuring that this sum is well-defined, provided the standard assumption of bounded rewards. \if\doctype 2
Thus, $v_\pi(s)$ is the expected discounted reward from starting in state $s$ and thereafter following policy $\pi$.
\fi
% \vspace{-5mm}\subsubsection*{Optimal Policies - }
The agent's goal is to learn an optimal policy, namely one that maximises the expected discounted return. This is actually equivalent to finding a policy that maximises the state-value function $v_\pi(s)$ at every state~\citep{sutton_bach_barto_2018}. 
%In fact, an optimal policy is defined as one that yields an optimal state-value function.
\begin{definition}
A policy $\pi$ is optimal if, at every state $s$, \if\doctype 2
$$
v_\pi(s) = v_*(s) = \max_{\pi'} v_{\pi'}(s).
$$
\else
$
v_\pi(s) = v_*(s) = \max_{\pi'} v_{\pi'}(s).
$
\fi
\end{definition}

%In theory one could consider non-stationary policies, which are policies $\pi(a|q,t)$ that may also depend on time $t$. However, because of the Markov property of MDPs which is that all relevant information to stochastic process that determines the next state is contained in the current state (in particular, past states are irrelevant), 
\if\doctype 2
It can be shown that, under standard assumptions, there will always be an optimal stationary (time independent) and deterministic (non-random) policy~\citep{watkins_1989}.
\fi
%Thus, typically we will search for an optimal stationary policy rather than considering non-stationary policies. 

\begin{definition}
Given a policy $\pi$, we can define a state-action-value function \if\doctype 2
$$v_\pi(s,a) = \E^\pi \left[ \left. \sum_{t=0}^\infty \gamma^t r(s_t,a_t) \right| s_0 = s, a_0 = a \right],$$
\else
$v_\pi(s,a) \allowbreak= \E^\pi\allowbreak \left[\allowbreak \left. \allowbreak\sum_{t=0}^\infty\allowbreak \gamma^t\allowbreak r(s_t,a_t) \right| s_0 = s, a_0 = a \right],$
\fi
similarly to the state-value function. This allows us to reinterpret the state-value function as \if\doctype 2
$$v_\pi(s) = \sum_a v_\pi(s,a) \pi(a|s),$$
\else
$v_\pi(s) = \sum_a v_\pi(s,a) \pi(a|s),$
\fi
and thus we can see that an optimal deterministic  policy $\pi$ must assign zero probability to any action $a$ that doesn't maximise the state-action value function. 
%In addition, given an optimal policy $\pi$, we can construct an optimal \textit{deterministic} policy $\pi_{det}$ by, for each state $q$, assigning probability 1 to some action $a$ maximising $v_\pi(q,a)$. 
\end{definition}
\if\doctype 2
We finally note that it is also possible to characterise optimal policies via the state-action-value function. Namely, 
$$
v_*(s,a) = \max_{\pi'} v_{\pi'}(s,a).
$$

\textbf{Off-policy RL}: Q-learning is a widely-used RL approach for learning an optimal policy. It is an \textit{off-policy} approach, which means that the final policy is independent from the action selection policy of the agent during learning. Q-learning works by constructing a Q-function $Q(s,a)$ which, through training, asymptotically converges to an optimal state-action-value function. A basic approach is \textit{one-step} Q-learning, in which the following update step is used: 
\begin{equation*}
    Q(s,a) \leftarrow (1-\mu)Q(s,a) + \mu\left(r(s,a) + \gamma 
    \max_{a'} Q(s',a')\right), 
\end{equation*}
where the constant $0<\mu\le1$ is known as the \textit{learning rate}. At the start of training, the values of $Q(s,a)$ are initialised to some arbitrary bounded values. 
%, often just to 0. 
At every step, after the agent selects action $a$ in state $s$ and moves to state $s'$, receiving reward $r(s,a)$, the update step above is performed. Under standard assumptions (on the learning rate and on persistent visitation on state-action pairs), the values $Q(s,a)$ converge to the optimal values $v_*(s,a)$~\citep{watkins_1989}. 
%First, there are some mild assumptions on the learning rate over time, which can always be fulfilled by choosing how the learning rate varies over time appropriately. More importantly, there is the assumption that every state-action pair is visited infinitely often. Many algorithms that employ Q-learning add a degree of randomness to their action-selection policy to ensure that this is the case. 
% In $\epsilon$-greedy Q-learning, the agent selects the action $a$ that has maximal Q-value with probability $1-\epsilon$, and with probability $\epsilon$ selects a random action - this ensures 
% %So long as the MDP is \textit{ergodic} in that every state is reachable from every other state, this small chance of random action selection ensures 
% that with probability all reachable state-action pairs are visited infinitely often. 
Once the Q-value has converged to the optimal values $v_*(s,a)$, the agent then has access to an optimal policy by simply choosing the action corresponding to the maximal $Q(s,a)$ in its current state.
\fi

\subsection{Dirichlet-Categorical Model }
%\label{sec:dirichlet}
%It is advantageous for an agent to construct and maintain a model of the MDP that can be use to reason about how to act safely. 
We consider a model for an MDP with unknown transition probabilities~\citep{ghavamzadeh_bayesian_2015}. The transition probabilities for a given state-action pair are assumed to be described by a categorical distribution over the next state. 
%Now, instead of maintaining a point-estimate of those transition probabilities, 
We maintain a Dirichlet distribution over the possible values of those transition probabilities: since the Dirichlet distribution is conjugate, we can employ Bayesian inference to update the Dirichlet distribution, as new observations are made while the agent explores the environment.    

Formally, for each state-action pair $(s^i,a)$, we have a Dirichlet distribution   $p^{i1}_a,p^{i2}_a,...,p^{iN}_a \sim Dir(\alpha^{i1}_a,\allowbreak\alpha^{i2}_a,...\alpha^{iN}_a)$, where $\mathbf{p^{i}_a}:=(p^{i1}_a,p^{i2}_a,...,p^{iN}_a)$, and the random variable $p^{ij}_a$ represents the agent's belief about the transition probability $P(s^j|s^i,a)$. At the start of learning, the agent will be assigned a prior Dirichlet distribution for each state-action pair, according to its initial belief about the transition probabilities. At every time step, as the agent moves from some state $s^i$ to some state $s^k$ by taking action $a$, it will generate a transition $s^i \xrightarrow{a} s^k$, which constitutes a new data point for the Bayesian inference. From Bayes' rule:
\begin{align*}
    &Pr(\mathbf{p^{i}_a} = \mathbf{q^{i}_a}|s^i \xrightarrow{a} s^k) \\
    &\propto Pr(s^i \xrightarrow{a} s^k|\bold{p^{i}_a} = \bold{q^{i}_a})Pr(\bold{p^{i}_a} = \bold{q^{i}_a}) \\
    & \quad = q^{ik}_a \prod_j (q^{ij}_a)^{\alpha^{ij}_a - 1} = [\prod_{j \ne k} (q^{ij}_a)^{\alpha^{ij}_a - 1}] (q^{ik}_a)^{(\alpha^{ik}_a + 1) - 1}, 
\end{align*}
where $\{q_a^{ij}\}_{j=1}^{N}$ belong to the standard $N-1$ simplex. This immediately yields 
\begin{equation*}
\resizebox{\columnwidth}{!}{$
    Pr(\mathbf{p^{i}_a} = \mathbf{q^{i}_a}|s^i \xrightarrow{a} s^k) = Dir(\alpha^{i1}_a,\alpha^{i2}_a,...,\alpha^{ik}_a +1,...,\alpha^{iN}_a).
    $}
\end{equation*} 
Thus, 
the posterior distribution is also a Dirichlet distribution. 
%which can be easily updated by simply adding 1 to the relevant parameter $\alpha^{ik}_a$. 
This update is repeated at each time step: 
the relevant information to the agent's posterior belief about the transition probabilities is the starting prior $Dir(\alpha^{i1}_a,\alpha^{i2}_a,...\alpha^{iN}_a)$ and the transition counts, keeping track of the number of times that $s^i \xrightarrow{a} s^j$ has occurred. The agent's posterior is then 
%$Dir(\alpha^{i1}_a + c^{i1}_a,...,\alpha^{iN}_a+c^{iN}_a)$. With the Dirichlet distribution 
$(p^{i1}_a,p^{i2}_a,...,p^{iN}_a) \sim Dir(\alpha^{i1}_a,\alpha^{i2}_a,...\alpha^{iN}_a)$: from this distribution, we can distill the expected value $\bar p^{ij}_a$ of each random variable $p^{ij}_a$, as well as the covariance of any two $p^{ij}_a$ and $p^{ik}_a$ (therefore also the variance of a single $p^{ij}_a$): 
\begin{align*}
    \bar p^{ij}_a = \E[p^{ij}_a] = \frac{\alpha^{ij}_a}{\alpha^{i0}_a},~ Cov[p^{ij}_a,p^{ik}_a] = \frac{\alpha^{ij}_a(\delta^{jk}\alpha^{i0}_a - \alpha^{ik}_a)}{(\alpha^{i0}_a)^2(\alpha^{i0}_a + 1)}, 
\end{align*}
where $\alpha^{i0}_a = \sum_{k=1}^N \alpha^{ik}_a$, and $\delta^{jk}$ is the Kronecker delta.

\section{Risk-aware Bayesian RL for Cautious Exploration}

In this section we propose a new approach to Safe RL, 
%that we will call \textit{Safe RL with Higher Order Information}
which will specifically address the problem of how to learn an optimal policy in an MDP with rewards while avoiding certain states classified as unsafe during training.
\if\doctype 2
The approach assumes that the MDP has unknown transition probabilities, and that the agent has a prior belief about those probabilities, which it updates as a Dirichlet-Categorical model, as described in the previous section. As a result, at any time the agent has access to the means $\bar p^{ij}_a := \E[p^{ij}_a]$ and the covariances $Cov[p^{ij}_a,p^{ik}_a]$ of its belief distributions  $p^{ij}_a$ about the MDP's transition probabilities $P(s^j|s^i,a)$. 

\fi
The agent is assumed to know which states of the MDP are safe and which are unsafe, but instead of assuming that the agent has this information globally, namely across all states of the MDP, we postulate that the agent observes states within an area around itself. This closely resembles real-world situations, where systems may have sensors that allow them to detect close-by dangerous areas, but not necessarily know about danger zones that are far away from them. In particular, we assume that there is an observation ``\radius'' $O$, such that the agent can observe all states that are reachable from the current state within $O$ steps and distinguish which of those states are safe or unsafe. The rest of this section is structured as follows: 

In Section~\ref{31}, we define the risk $\rho^m(s,a)$ over $m$ steps of taking an action $a$ at the current state~$s$. We then introduce a random variable $\varrho^m(s,a)$ representing the agent's belief about the risk; 
%\textcolor{red}{[why is index 'c' used here?]}~\edit{[Index 'c' is indeed redundant but I believe it is referring to the 'c'urrent state, and might be good to keep it.]}
In Section~\ref{32}, we leverage a method from~\citet{casella_berger_2021} to approximate the expected value and variance of the random variable $\varrho^m(s,a)$. We provide convergence results on the approximations of the expectation and variance of $\varrho^m(s,a)$; In Section~\ref{sec:confidence}, we show how the Cantelli Inequality \citep{cantelli1929sui} allows us to estimate a confidence bound on the risk $\rho^m(s,a)$; In Section~\ref{34}, we prescribe a methodology for incorporating the expectation and variance of the risk into the local action selection during the training of the RL agent.

\subsection{Definition and Characterisation of the Risk}\label{31} 

Given the observation \radius $O$, we reason about the risk incurred over the next $m$ steps after taking a particular action $a$ in the current state $s$, for any $m \le O$. 
%For a particular value of the horizon, $m$, one might like to define the $m$-step risk in a state $s$ of an action $a$ as the probability of entering an unsafe state within $m$ steps after selecting that action. 
However, note that there is a dependence between the agent's estimate of such a risk and the use of that estimate to inform its action selection policy. 
%, calculating a risk defined in that way would be infeasible since the risk depends on the agent's action selection policy, which in turn depends on the risk. 
In order to solve this dilemma we fix a policy over the $m$-step horizon, and calculate the corresponding risk, given that policy. Similar to temporal-difference learning schemes, this is done by assuming best-case action selection, namely, the $m$-step risk $\rho^m(s,a)$ at state $s$ after taking action $a$ is defined assuming that after selecting action $a$, the agent will select subsequent actions to minimize the expected risk. 
Assuming that the agent is at state $s$, we define the agent's approximation of the $m$-step risk $\bar \varrho^{~m}(s,a)$ by back-propagating the risk given the ``expected safest policy'' over $m$ steps, as follows: 
\begin{align}
\bar \varrho^{~n+1}(s^k,a) & =  
    \begin{cases}\label{eq:risk_backprop}
      1 & \text{$s^k$ observed and unsafe}\\
      \sum\limits^{N}_{j=1} \bar p^{kj}_a \bar \varrho^{~n}(s^j) & \text{otherwise};
    \end{cases} \\
\bar \varrho^{~n}(s^k) & :=  
    \begin{cases}
      1 & \text{$s^k$ observed and unsafe}\\
      \min\limits_{a\in A} \bar \varrho^{~n}(s^k, a) & \text{otherwise};
    \end{cases} \\
\bar \varrho^{~0}(s^k) & :=  \mathds{1}(s^k \text{ is observed and unsafe}).\label{eq:safe_id}
\end{align}  
We terminate this iterative process at $n+1=m$ and once we have calculated $\bar \varrho^{~m}(s,a)$, for actions $a \in A$. Note that, despite the use of progressing indices $n$, this is an iterative back-propagation that leverages the expected values of agent's belief about the transition probabilities, i.e., $\bar p^{kj}_a$. Thus, $\bar \varrho^{~m}(s,a)$ is the agent's approximation of the expectation of the probability of entering an unsafe state within $m$ steps by selecting action $a$ at state $s$, and thereafter by selecting actions that it currently believes will minimize the probability of entering unsafe states over the given time horizon. 
%\edit{If needed, we can remove this sentence:} Note that in \eqref{eq:risk_backprop}, the term $\bar R^m_k(a)$ for another state $s^k$ is \textit{not} the agent's approximation of the probability of entering an unsafe state within $m$ steps by selecting action $a$ in the state $s^k$, since the set of states that are observed is relative to the agent's current state $s$. 

\begin{remark}
    We note that, in practice, an autonomous agent can determine, with some certainty, whether a subset of its observation are is safe to visit or not. Consider a mobile robot that moves in an office environment and can deem certain states as obstacles-to-avoid based on the received signals from onboard sensors. It is straightforward to extend the indicator function in \eqref{eq:safe_id} to a probability distribution, to reflect agent uncertainty over such signals. 
\end{remark}

The term $\bar p^{kj}_a = \E[p^{kj}_a]$ is used as a point estimate of the true transition probability $t^{kj}_a = P(s^j|s^k,a)$. 
The value of $\bar \varrho^{~m}(s,a)$ only relies on states which the agent believes are reachable from $s$ within $m$ steps. In particular so long as the horizon $m$ is less than the observation \radius $O$, the agent is able to observe all states which are relevant to the calculation of $\bar \varrho^{~m}(s,a)$, so specifically, $\mathds{1}(s^j \text{ is unsafe}) = \mathds{1}(s^j \text{ is observed and unsafe})$ for all relevant states $s^j$ (see Appendix \ref{appndix:approximation} for more details).

\subsection{Approximation of Expected Value and Covariance of the Risk}\label{32}
In the previous section, we presented the underlying mechanism for calculating an $m$-step \textit{expected} risk. However, relying only on this expected value disregards the agent's confidence placed over this expectation: as a shortcoming of this, the agent might be willing to take actions that have lower expected risk, but which come with lower confidence as well.  Evidently this behavior can be unsafe, and we would prefer the agent to employ its confidence in the decision-making process. In the following, we formalize the underpinnings of how to incorporate a confidence approximation into the agent action selection policy.

Let $\mathbf{x}$ denote the vector of variables $x^{ij}_a$ where $i,j$ range from $1$ to $N$ and $a$ ranges over $A$, i.e., $\mathbf{x} =\left((x^{ij}_a)_{i,j=1,...,N \text{ and } \forall a \in A}\right)$. We assume that these indices are ordered lexicographically by $(i,a,j)$. This is because $i$ and $a$ will be used to signify a state-action pair $(s^i,a)$, and $j$ will be used to signify a potential next state $s^j$. Introduce a set of 
%\edit{``risk polynomials'' [not super clear why poly?]} \edit{[ROHAN: I've just called these this because it makes it easier to refer back to them later. They just happen to be polynomials because bellman back-propagated risk is polynomial in the transition probabilities. Maybe a different name is more appropriate?]} 
functions $g^n_k[\mathbf{x}]$ (we shall see they take the shape of polynomials), defined as follows for each state $s^k$:
\begin{align*}
\begin{aligned}
  &g^{n+1}(s^k,a)[\mathbf{x}]  :=
    \begin{cases}
      1 & \text{\hspace{-22mm}if $s^k$ is observed and unsafe}\\
      \sum\limits^N_{j=1} x^{kj}_a g^{n}(s^j)[\mathbf{x}] & \text{otherwise};
    \end{cases} \\
  &g^{n}(s^k)[\mathbf{x}]  :=
    \begin{cases}
      1 & \text{\hspace{-39mm}if $s^k$ is observed and unsafe}\\
      g^{n} \left(s^k, \argmin\limits_a \bar \varrho^{~n}_{k}(a) \right)[\mathbf{x}]  & \text{otherwise};
    \end{cases} \\
&g^0(s^k)[\mathbf{x}]  := \mathds{1}( s^k \text{ is observed and unsafe}).
\end{aligned}
\end{align*}
Then we can write the risk (of selecting action $a$ in state~$s$, over $m$ steps) defined above as $\rho^m(s,a) = g^m(s,a)[\mathbf{t}]$, where $\mathbf{t} =\left((t^{ij}_a)_{i,j=1,...,N \text{ and } \forall a \in A}\right)$ is a vector of all ``true'' transition probabilities, namely $t^{ij}_a = P(s^j|s^i,a)$. We can similarly write the agent's approximation of the expected risk, as described in Section~\ref{31}, as $\bar \varrho^{~m}(s,a) = g^m(s,a)[\bar{\mathbf{p}}]$, where similarly $\bar{\mathbf{p}} =\left((\bar p^{ij}_a)_{i,j=1,...,N \text{ and } a \in A}\right)$, and $\bar p^{ij}_a$ is the expected value of each random variable $p^{ij}_a$. We refer to the actions specified by the $\argmin$ operators as the \textit{agent's expected safest action} in each state over the next $m$ steps.

Now, crucially, we can also define a new random variable $\varrho^m(s,a) = g^m(s,a)[\mathbf{p}]$, where $\mathbf{p} =\left(( p^{ij}_a)_{i,j=1,...,N \text{ and } \forall a \in A}\right)$. Since the $p^{ij}_a$s are random variables representing the agent's beliefs about the true transition probabilities $t^{ij}_a$, we in fact have that this random variable $\varrho^m(s,a)$ represents the agent's beliefs about the true risk $\rho^m(s,a)$. In the following, we show that $\bar \varrho^m(s,a)$ can be viewed as an approximation of $\E [\varrho^m(s,a)]$, and we provide and justify an approximation of $\textit{Var} [\varrho^m(s,a)]$ that is directly correlated to agent's confidence on $\E [\varrho^m(s,a)]$. These approximations can be used by the agent to reason more accurately about the true risk of selecting an action $a$ in a state $s$, over $m$ steps, i.e., $r^m(s,a)$.

In order to construct approximations of the expectation and the variance of $R^m(s,a)$, we make use of the first-order Taylor expansion of $g^m(s,a)[\mathbf{x}]$ around $\mathbf{x} = \mathbf{\bar p}$, following a method in \citep{casella_berger_2021}. The first-order Taylor expansion is 
\begin{equation*}\label{eq:taylor_expansion}
\resizebox{\columnwidth}{!}{$
    g^m(s,a)\left[\mathbf{x}\right] = g^m(s,a)\left[\mathbf{\bar p}\right] + \sum_{i,j=1}^N \sum_{b \in A} \frac{\partial g^m(s,a)}{\partial x^{ij}_b} (x^{ij}_b - \bar p^{ij}_b),% + \text{ remainder,}
    $}
\end{equation*}  
where the partial derivatives are also evaluated at $\mathbf{\bar p}$ and we have disregarded the remainder term. Reasoning over the random variables $\mathbf{p}$ for $\mathbf{x}$:
\begin{equation}\label{eq:statistical_approximation}
\resizebox{0.9\columnwidth}{!}{$
    g^m(s,a)\left[\mathbf{p} \right] \approx g^m(s,a)\left[\mathbf{\bar p}\right] + \sum_{i,j=1}^N \sum_{b \in A} \frac{\partial g^m(s,a)}{\partial x^{ij}_b} (p^{ij}_b - \bar p^{ij}_b).
    $}
\end{equation}
We can then take the expectation of both sides, obtaining  
\begin{align}
    &\E[ g^m(s,a)\left[\mathbf{p} \right]]\\
    &\approx \E[ g^m(s,a)\left[\mathbf{\bar p}\right]] + \E[ \sum_{i,j=1}^N \sum_{b \in A} \frac{\partial g^m(s,a)}{\partial x^{ij}_b} (p^{ij}_b - \bar p^{ij}_b)] \nonumber \\
    & = g^m(s,a)\left[\mathbf{\bar p}\right] + \sum_{i,j=1}^N \sum_{b \in A} \frac{\partial g^m(s,a)}{\partial x^{ij}_b} \E[ (p^{ij}_b - \bar p^{ij}_b)] \nonumber  \\
    & = g^m(s,a)\left[\mathbf{\bar p}\right],
\end{align}
where the above steps follow since the only random term in the right-hand side is $p^{ij}_b$, for which $\E (p^{ij}_b) = \bar p^{ij}_b$. Also, recall that $g^m(s,a)\left[\mathbf{p} \right] = \varrho^m(s,a)$ and $g^m(s,a)\left[\mathbf{ \bar p} \right] = \bar \varrho^{~m}(s,a)$. Thus, we have $\bar \varrho^m(s,a)$ as an approximation of the expectation of $\varrho^m(s,a)$. For the approximation of the variance of the agent's believed risk, which is again a random variable, we can write: 
\begin{align}
    &\textit{Var}(g^m(s,a)[\mathbf{p}])\\
    &\approx \E [(g^m(s,a)[\mathbf{p}] -  g^m(s,a)[\mathbf{\bar p}])^2] \nonumber\\
    &\approx \E\left[ \left(\sum_{i,j=1}^N \sum_{b \in A} \frac{\partial g^m(s,a)}{\partial x^{ij}_b} (p^{ij}_b - \bar p^{ij}_b)\right)^2\right] \tag{from \eqref{eq:statistical_approximation}} \\
    &= \sum_{i,j,s,t=1}^N \sum_{b_1,b_2 \in A} \frac{\partial g^m(s,a)}{\partial x^{ij}_{b_1}} \frac{\partial g^m(s,a)}{\partial x^{st}_{b_2}} \textit{Cov}(p^{ij}_{b_1},p^{st}_{b_2}) \nonumber\\
    &= \sum_{i = 1}^N \sum_{b \in A} \sum_{j,t=1}^N \frac{\partial g^m(s,a)}{\partial x^{ij}_b} \frac{\partial g^m(s,a)}{\partial x^{it}_b} \textit{Cov}(p^{ij}_b,p^{it}_b) \\
    &:= \bar V^m(s,a),\label{eq:variance_approximation}
\end{align}
where $\bar V^m(s,a)$ is the approximation for the variance of $\varrho^m(s,a)$, i.e., $\bar V^m(s,a) \approx \textit{Var}(\varrho^m(s,a))$, and the last line follows from the fact that the covariance between two transition probability beliefs $p^{ij}_{b_1}$ and $p^{st}_{b_2}$ is always $0$, unless they correspond to the same starting state-action pair $(s^i,b)$. In other words, $\textit{Cov}(p^{ij}_{b_1},p^{st}_{b_2}) = 0$ unless $i = j$ and $b_1 = b_2$. Next, we show consistency of the estimate in the limit (see Appendix \ref{appndix:convergence_results} for the proof). 
\begin{theorem}
\label{thm:convergence}
Under standard Q-learning convergence assumptions~\citep{watkins_1989}, namely that reachable state-action pairs are visited infinitely often, the estimate of the mean of the believed risk distribution $\bar \varrho^m(s,a)$ converges to the true risk $\rho^m(s,a)$, and it does so with the variance of the believed risk distribution $\textit{Var}\left(g^m(s,a)[\mathbf{p}]\right)$ approaching the estimate of that variance $\bar V^m(s,a)$. Specifically, 
\begin{equation*}
    \frac{\bar \varrho^m(s,a) - \rho^m(s,a)}{\sqrt{\bar V^m(s,a)}} \rightarrow \mathcal{N}(0, 1) \text{ in distribution. }
\end{equation*}
\end{theorem}

%\proof (see Appendix \ref{appndix:convergence_results})

\subsection{Estimating a Confidence on the Approximation of the Risk}\label{sec:confidence}

So far we have shown that when the agent is in the state $s$, for each possible action $a$, approximations of the expectation and variance of its belief $\varrho^m(s,a)$ about the risk $\rho^m(s,a)$ can be formally obtained: we have denoted these two approximations by $\bar \varrho^m(s,a)$ and $\bar V^m(s,a)$, respectively. We now describe a method for combining these approximations to obtain a bound on the level of confidence that the risk $\rho^m(s,a)$ is below a certain threshold.  

We appeal to the Cantelli Inequality, which is a one-sided Chebychev bound~\citep{cantelli1929sui}. Having computed $\bar \varrho^m(s,a)$ and $\bar V^m(s,a)$, for a particular confidence value $0<C<1$ we can define $\Phi := \bar \varrho^m(s,a) + \sqrt{\frac{\bar V^m(s,a)C}{1-C}}$. From the Cantelli Inequality we then have
\begin{equation*}
    \Pr(\varrho^m(s,a) \le \Phi ) \ge C. 
\end{equation*}
Specifically, $\Phi$ is the lowest risk level such that, according to its approximations, the agent can be at least $100\times C~\%$ confident that the true risk is below level $\Phi$. The exploration mechanism can therefore leverage $\Phi$ to ensure that the required safety level is met (please refer to Appendix \ref{appndix:confidence_bound} for more details).

\subsection{RCRL: Risk-aware Bayesian RL for Cautious Exploration}\label{34}

In this section we propose an overall approach for safe RL, which leverages the expectation and variance of the defined risk measure to allow an agent to explore the environment safely, while attempting to learn an optimal policy. 
In order to select an optimal-yet-safe
action at each state, we propose a \emph{double-learner} architecture, 
referred to as \textit{Risk-aware Cautious RL (RCRL)} and explained next. 

% The Q-learning part of our agent  will simply perform one-step Q-learning as in section 2.3, starting with an arbitrary finite Q function $Q(s,a)$ for every state-action pair, and updating it following 
% \begin{equation}
%     Q(q,a) \leftarrow (1-\mu)Q(q,a) + \mu\left(re(q,a) + \gamma 
%     \max_{a'} Q(q',a')\right) 
% \end{equation}

% after every time a given state-action pair $(q,a)$ is visited. This is the part of the agent that will attempt to learn a policy that maximises rewards. 

% We must now determine an action-selection scheme to compliment Q-learning that will keep the agent safe during exploration. 

The first learner is an optimistic agent whose objective is to maximize the expected cumulative return\if\doctype 2
, as described in Section \ref{sec:QL}
\fi
. 
The second learner is a pessimistic agent that maintains a Dirichlet-Categorical model of the transition probabilities of the MDP. In particular, this agent is initialized with a prior that encodes any information the agent might have about the transition probabilities. For each state-action pair $(s^i,a)$ we have a Dirichlet distribution $p^{i1}_a,p^{i2}_a,...,p^{iN}_a \sim Dir(\alpha^{i1}_a,\alpha^{i2}_a,...\alpha^{iN}_a)$. As the agent explores the environment, the Dirichlet distributions are updated using Bayesian inference.

For each action $a$ available in the current state $s$, the pessimistic learner computes the approximations $\bar \varrho^m(s,a)$ and $\bar V^m(s,a)$ of its belief $\varrho^m(s,a)$ of the risk, over the next $m$ steps, associated to taking action $a$ in $s$. The ``risk horizon'' $m$ is a hyper-parameter that, as discussed, should be set to be at most the observation \radius $O$. 
The pessimistic learner is initialized with two extra hyper-parameters $\Phi_{max}$ and $C(n)$: $\Phi_{max}$ represents the maximum level of risk that the agent should be prepared to take, whereas $C(n)$ is a decreasing function of the number of times $n$ that the current state has been visited, which satisfies $C(0)<1$ and $\lim_{n \rightarrow \infty} C(n) = 0$. From Section \ref{sec:confidence}, the agent can then
%, as in section 4.4, use 
compute, for each action $a$, the value 
\begin{equation}
\label{eqn:PaDefinition}
    \Phi = \bar \varrho^m(s,a) + \sqrt{\frac{\bar V^m(s,a)C(n)}{1-C(n)}},
\end{equation}
which can thus define a set of safe actions: these are all the actions that the agent believes have risk less than $\Phi_{max}$, with confidence at least $C(n)$, namely   
\begin{equation*}
    A_{\textit{safe}} = \{a \in A | \Phi \le \Phi_{max} \}.  
\end{equation*}
In case there are no actions $a$ such that $\Phi \le \Phi_{max}$, the agent instead allows  
\begin{equation}\label{eq:safety_mode}
    A_{\textit{safe}} = \{ a \in A | \bar \varrho^m(s,a) = \min_{a'} \bar \varrho^m(s,a') \}. 
\end{equation}

Finally, the agent selects an action $a^*_{\textit{safe}}$ from the set of safe actions according to the Q-values of those actions, e.g., using softmax action selection~\citep{sutton_bach_barto_2018} with some \textit{temperature} $\mathcal{T}>0$: 
\begin{equation}
\label{eq:boltzmann-rational}
    Pr( a^*_{\textit{safe}} = a) = \frac{e^{Q(s,a)/\mathcal{T}}}{\sum_{a \in A_{\textit{safe}}} e^{Q(s,a)/\mathcal{T}} }. 
\end{equation}

The pseudo-code for the full algorithm is presented in Algorithm~1.

\begin{algorithm}[t]
\SetAlgoLined
\SetNlSty{texttt}{(}{)}
\SetKwInOut{Input}{input}
\caption{Risk-aware Cautious RL (RCRL)}
\Input{$\textit{Prior}$, $C(n)$, $\Phi_{max}$, $max\_steps$, $max\_episodes$, $\mu$, $\gamma$, $m$ }
\BlankLine
\nl initialize $Q(s,a)$ for each state-action pair $(s,a)$\;
\nl initialize $num\_steps = 0$ \;
\nl initialize $num\_episodes = 0$ \;
\While{$num\_episodes < max\_episodes$}{
\nl $s \leftarrow s^0$\;
\nl $num\_episodes \leftarrow num\_episodes + 1$\;
\While{$num\_steps < max\_steps$ and $s$ is not unsafe}{
\nl calculate $\bar \varrho^m(s,a)$ as in (\ref{eq:risk_backprop}) \;
\nl calculate $\bar V^m(s,a)$ as in (\ref{eq:variance_approximation}) \;
\nl calculate $\Phi$ as in (\ref{eqn:PaDefinition}) \;
\nl $A_{\textit{safe}} := \{a \in A | \Phi \le \Phi_{max} \}$ \;
\If{$A_{\textit{safe}} = \emptyset$}{
\nl $A_{\textit{safe}} \leftarrow \{ a \in A | \bar \varrho^m(s,a) = \min_{a'} \bar \varrho^m(s,a') \}$ \;}
\nl choose action $a^*_{\textit{safe}}$ according to (\ref{eq:boltzmann-rational}) \;
\nl execute action $a^*_{\textit{safe}}$ to environment and receive next state $s'$ and reward $r(s,a^*_{\textit{safe}})$ \;
\nl update belief $p$ as in section \ref{sec:QL} \;
\nl update $Q(s,a^*_{\textit{safe}}) \leftarrow (1-\mu)Q(s,a^*_{\textit{safe}}) + \mu\left(r(s,a^*_{\textit{safe}}) + \gamma 
    \max_{a'} Q(s',a')\right)$ \;
\nl $s \leftarrow s'$ \;
\nl $num\_steps \leftarrow num\_steps + 1$\;
}
}
\label{algo}
\end{algorithm}

\begin{remark}\label{rem:safetyVSopt}
It is worth emphasizing that the main objective of RCRL is to maintain safety for cautious exploration, for which theoretical guarantees are provided. The optimal policy in the sense of traditional RL, i.e. maximising the reward return, might not be safe in many scenarios. While RCRL pushes the exploration boundaries as the agent becomes more certain about the risk, the main priority in RCRL is to maintain the agent safety and not to maximize the expected reward.
\end{remark}

In summary, we effectively have two agents learning to accomplish two tasks. The first agent performs Q-learning to learn an optimal policy for the reward. The second agent determines the best approximation of the expected value and variance of each action, enabling it to prevent  the first agent from selecting actions that it cannot guarantee to be safe enough (with at least a given confidence). 
When instead the pessimistic agent cannot guarantee that any action is safe enough, it forces the optimistic learner to go into ``safety mode'', i.e., to forcibly select the actions that minimize the expected value of the risk, as per \eqref{eq:safety_mode}. 
From an empirical perspective, implementing this concept of a ``safety mode'' allows for continued progress, and pairs well with the definition of risk: namely, when the agent deems that a state is too risky, it will go into this ``safety mode'' until it is back in a state with sufficiently safe actions. 

Finally, note that $C(n)$ represents the level of confidence that the agent requires in an action being safe enough for it to consider taking that action. When the agent starts exploring and $C(n)$ is at its highest, the agent only explores actions that it is very confident in. However, it may need to take actions that it is less confident in order to find an optimal policy. Thus, as it continues exploring, $C(n)$ is reduced, allowing the agent to select actions upon which it is not as confident. However, in the limit, when $C(n) \rightarrow 0$, we have that $\Phi = \bar \varrho^m(s,a)$, which means that the agent never takes an action if its approximation of the expected risk $\bar \varrho^m(s,a)$ is more than the maximum allowable risk $\Phi_{max}$.  

%Thus, if $\Phi_{max}$ is set too low there may be some optimal action $a$ which is never taken by the agent, and therefore the Q-learning may not converge to the optimal policy. Nevertheless, in many cases the actions judged with $\bar R^m_c(a) > \Phi_{max}$ may all be sub-optimal due to the associated risk, and in such cases Q-learning will still converge to the optimal policy.
%

\newcommand{\x}{0.45}
\begin{figure}[!ht]
\centering
    \begin{subfigure}[b]{\x\linewidth}
    \centering
    \includegraphics[width=0.99\linewidth]{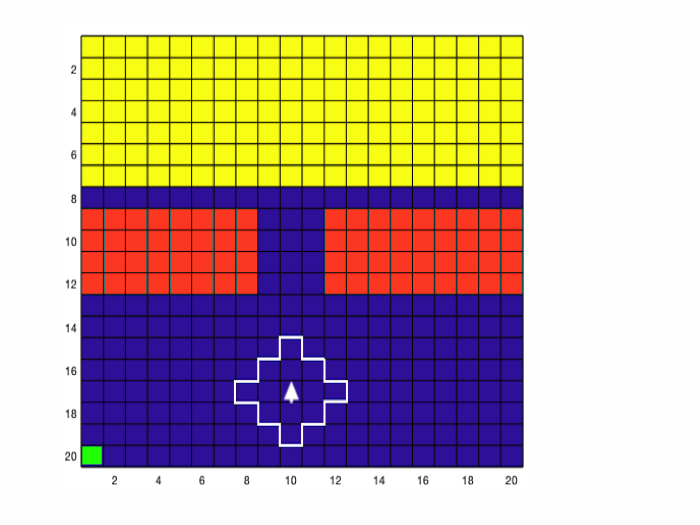} 
    \caption{} 
    \label{fig:MDP_states} 
    \vspace{1ex}
  \end{subfigure}%% 
  \begin{subfigure}[b]{\x\linewidth}
    \centering
    \includegraphics[width=1\linewidth]{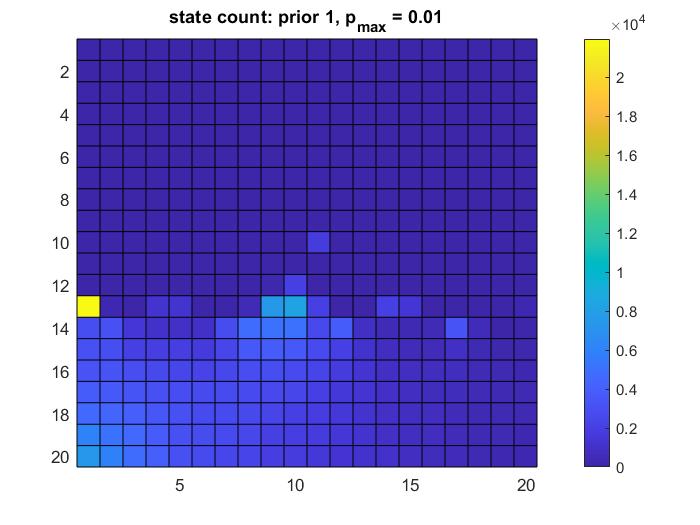} 
    \caption{} 
    \label{state_count_uninformative}
    \vspace{1ex}
  \end{subfigure}
  \begin{subfigure}[b]{\x\linewidth}
    \centering
    \includegraphics[width=1\linewidth]{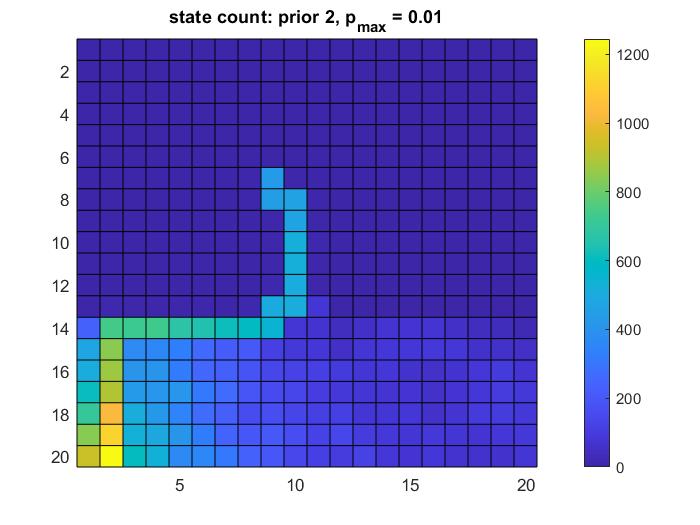} 
    \caption{} 
    \label{fig7:a} 
    \vspace{1ex}
  \end{subfigure}%% 
  \begin{subfigure}[b]{\x\linewidth}
    \centering
    \includegraphics[width=1\linewidth]{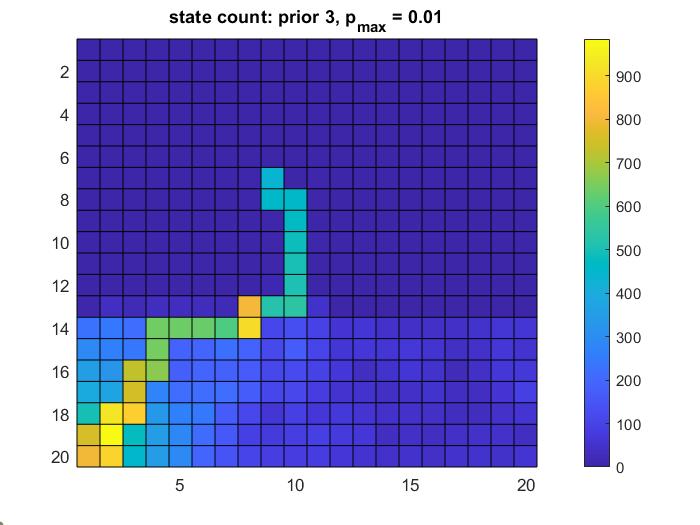} 
    \caption{} 
    \label{fig7:b} 
    \vspace{1ex}
  \end{subfigure} 
  \caption{(a) Slippery BridgeCross setup: agent is represented by an arrow surrounded by the observation area (white line). Labels denote target (yellow), unsafe (red) and safe states (blue), and  initial state ($s_0$, green). (b) For a single experiment, number of state-visitations for Prior 1 at $\Phi_{max}=0.01$. (c-d) Number of state-visitations, for Priors 2 and 3 at $\Phi_{max}=0.01$.}
\end{figure}

\section{Experiments}

\textbf{BridgeCross - } We first evaluated the performance of RCRL on a \emph{Slippery Bridge Crossing} example. The states of the MDP consist of a $20 \times 20$-grid, as depicted in Figure \ref{fig:MDP_states}. The agent is initialized at $q_0$ in the bottom-left corner (green). 
The agent's task is to get to the goal region without ever entering an unsafe state. In particular, upon reaching a goal state, the agent is given a reward of 1 and the learning episode is terminated; at every other state it receives a reward of 0, and upon reaching an unsafe state the learning episode  terminates with reward 0. 
%For the rest of this chapter we will refer to entering an unsafe state as ``sinking'', with the idea that the agent attempts to cross a bridge without falling into water on either side. 

\begin{table}[!t]
\caption{Number of successes and failures over total episodes. BridgeCross: different priors and acceptable risks $\Phi_{max}$, averaged over 10 agents.  Pacman: varying risk horizon $m$, single agent.}\label{tab:tab1}
\centering
\resizebox{\columnwidth}{!}{
\begin{tabular}{|c|c|c|c|c|c|c|}
\hline
Experiment                     & $|S|$ & $|A|$ & Safety Setup                 & \# Successes & \# Failures & Total Episodes \\ \hline
\multirow{14}{*}{BridgeCross} & 400   & 5     & Prior 1, $\Phi_{max}=0.33$   & 404.3       & 54.2       & 500            \\ \cline{2-7} 
                               & 400   & 5     & Prior 1, $\Phi_{max}=0.01$   & 506.0       & 417.9      & 1500           \\ \cline{2-7} 
                               & 400   & 5     & Prior 2, $\Phi_{max}=0.33$   & 424.3       & 32.1       & 500            \\ \cline{2-7} 
                               & 400   & 5     & Prior 2, $\Phi_{max}=0.01$   & 384.6       & 0.5        & 500            \\ \cline{2-7} 
                               & 400   & 5     & Prior 3, $\Phi_{max}=0.01$   & 407.4       & 14.4       & 500            \\ \cline{2-7} 
                               & 400   & 5     & Prior 3, $\Phi_{max}=0.0033$ & 421.3       & 1.1        & 500            \\ \cline{2-7} 
                               & 400   & 5     & QL with Penalty              & 414.6       & 990.5      & 1500           \\ \cline{2-7} 
                               & 400   & 9     & Prior 1, $\Phi_{max}=0.33$   & 299.1       & 173.4      & 500            \\ \cline{2-7} 
                               & 400   & 9     & Prior 1, $\Phi_{max}=0.01$   & 348.9       & 523.2      & 1500           \\ \cline{2-7} 
                               & 400   & 9     & Prior 2, $\Phi_{max}=0.33$   & 444.7       & 38.9       & 500            \\ \cline{2-7} 
                               & 400   & 9     & Prior 2, $\Phi_{max}=0.01$   & 17.6        & 14.5       & 500            \\ \cline{2-7} 
                               & 400   & 9     & Prior 3, $\Phi_{max}=0.01$   & 391.7       & 15.4       & 500            \\ \cline{2-7} 
                               & 400   & 9     & Prior 3, $\Phi_{max}=0.0033$ & 430.0       & 2.2        & 500            \\ \cline{2-7} 
                               & 400   & 9     & QL with Penalty              & 367.9       & 1119.2     & 1500           \\ \hline
\multirow{3}{*}{Pacman}        & 4000  & 5     & Risk Horizon $m = 2$         & 234         & 77         & 311            \\ \cline{2-7} 
                               & 4000  & 5     & Risk Horizon $m = 3$         & 207         & 68         & 275            \\ \cline{2-7} 
                               & 4000  & 5     & QL with Penalty              & 0           & 1500       & 1500           \\ \hline
\end{tabular}}
\label{tab:average_totals}
\vspace*{-1\baselineskip}
\end{table}

% \begin{table*}[!t]
% \caption{Number of successes and failures over total episodes. BridgeCross: different priors and acceptable risks $\Phi_{max}$, averaged over 10 agents.  Pacman: varying risk horizon $m$, single agent.}\label{tab:tab1}
% \centering
% \begin{tabular}{| c |c|c|l || r | r | r|}
% \hline
% Experiment & $|S|$ & $|A|$ & Safety Setup & \# Successes & \# Failures & Total Episodes\\
% \hline \hline
% \multirow{6}{*}{
% BridgeCross
% } & 400 & 5 & Prior 1, $\Phi_{max} = 0.33$ &  404.3 & 54.2 & 500 \\
% \cline{2-5}
% & 400 & 5 & Prior 1, $\Phi_{max} = 0.01$ & 506.0 & 417.9 & 1500 \\
% \cline{2-5}
% & 400 & 5 & Prior 2, $\Phi_{max} = 0.01$ &  384.6 & 0.5 & 500 \\
% \cline{2-5}
% & 400 & 5 & Prior 3, $\Phi_{max} = 0.01$ & 407.4 & 14.4 & 500 \\
% \cline{2-5}
% & 400 & 5 & Prior 3, $\Phi_{max} = 0.0033$ & 421.3 & 1.1 & 500 \\
% \cline{2-5}
% & 400 & 5 & QL with Penalty & 414.6 & 990.5 & 1500 \\
% \hline \hline
% \multirow{3}{*}{
% Pacman
% } & 4000 & 5 & Risk Horizon $m = 2$ &  234 & 77 & 311 \\
% \cline{2-5}
% & 4000 & 5 & Risk Horizon $m = 3$ &  207 & 68 & 275 \\
% \cline{2-5}
% & 4000 & 5 & QL with Penalty & 0 & 1500 & 1500 \\
% \hline
% \end{tabular}
% \label{tab:average_totals}
% \vspace*{-1\baselineskip}
% \end{table*}

\textcolor{black}{We consider two cases regarding the action space. Case~I: We assume that}  
at each time step the agent might move into one of the 4 neighbouring states, or stay in its current position; thus, the agent has access to 5 actions at each state, $A = \{\mathit{right}, \mathit{up}, \mathit{left}, \mathit{down}, \mathit{stay}\}$. 
\textcolor{black}{Case~II: We consider a larger action space that includes the diagonal actions as well, i.e., $|\mathcal{A}|=9$. In both cases,}
if the agent selects action $a \in A$, then it has a $96\%$ chance of moving in direction $a$, and a $4\%$ chance of ``slipping'', namely moving into another random direction. 
%(including staying still). 
If any movement would ever take the agent outside of the map, then the agent will just remain in place. The agent is assumed to have an observation \radius $O=2$ steps. Note that due to the slipperiness of the movement and the narrow passage to reach the goal state, minimizing the risk is not aligned with maximizing the expected reward.

We tested RCRL with 5 different combinations of a prior and a maximum acceptable risk $\Phi_{max}$. The following additional hyper-parameters of the algorithm were kept constant: the maximum number of steps per episode $max\_steps = 400$, the maximum number of episodes $max\_episodes = 500$ (although this was increased to 1500 in two cases when the agent did not converge to  near-optimal policy within the first $500$, cf. Table \ref{tab:average_totals}); the learning rate $\mu = 0.85$; the discount factor $\gamma = 0.9$; and the risk horizon $m = 2$ (Algorithm 1). 
%which was set to match the observation \radius.
Recall that a prior consists of a Dirichlet distribution   $p^{i1}_a,...,p^{iN}_a \sim \allowbreak Dir(\alpha^{i1}_a,..\allowbreak.,\alpha^{iN}_a)$ for every state-action pair $(s^i,a)$. We considered three priors: 
\begin{itemize}
    \item Prior 1 - completely uninformative: in this case we assigned a value of 1 to every $\alpha$. This yields a distribution that is uniform over its support. 
    \item Prior 2 - weakly informative: we assigned a value of 12 to the $\alpha$ corresponding to moving in the correct direction, and a value of 1 to all other $\alpha$'s. This gives a distribution in between Prior 1 and Prior 3 in both degree of bias and concentration.
    \item Prior 3 - highly informative: we assigned a value of 96 to the $\alpha$ corresponding to moving in the correct direction, and a value of 1 to all other $\alpha$'s. This gives a distribution that is highly concentrated, and for which the mean values of the transition probability random variables are the true transition probabilities of the MDP, and hence unbiased. 
\end{itemize}

We tested the algorithm with all three priors and a maximum acceptable risk of $\Phi_{max} = 0.01$ and repeating each experiment 10 times to take averages. %\textcolor{blue}{The results are summarized in Table \ref{tab:average_totals}.}
We first discuss the results for Case I. On average, the agent with the highly informative prior (Prior 3) entered unsafe states 14.4 times (on average), and always converged to near-optimality within about 200 steps, successfully crossing the bridge 407.4 times. For the other 78.2 episodes, the agent reached the episode limit before crossing the bridge or entering an unsafe state. The agent with Prior 2 interestingly only entered unsafe states an average of 0.5 times per experiment, and converged to a near-optimal policy within about 300 episodes, successfully crossing the bridge 384.6 times. On the other hand, the agent with Prior 1 only crossed the bridge less than 30 times. We therefore increased the total number of episodes to 1500 and tried again, yet still over half the time it did not converge to a near-optimal policy (Figure~\ref{average_steps_to_win}). \textcolor{black}{A similar pattern is observed for Case II, where the number of failed episodes tends to decrease as the prior becomes more informative. Interestingly, the agent with Prior 2 also exhibits a relatively low number of successful episodes. A potential explanation for this could be the low acceptable risk of $\Phi_{\text{max}}=0.01$, as discussed in Remark \ref{rem:safetyVSopt}. It's noteworthy that augmenting the action space does not necessarily lead to an improvement in the number of successful or failed episodes. As anticipated, the increase in the number of actions corresponds to an expected increase in both the total training runtime and the runtime for computing the set of safe action (Figure~\ref{fig:runtimes}).}

We then tested \textcolor{black}{Prior 1/Case I} with a more lenient maximum acceptable risk of $\Phi_{max} = 0.33$, and found that the agent this time managed to converge to near-optimality within around 200 episodes, entering unsafe states 54.2 times and successfully crossing the bridge 404.3 times. We also tested \textcolor{black}{Prior 3/Case I} with a stricter $\Phi_{max} = 0.0033$ and found out that it entered unsafe states only 1.1 times and succeeded 421.3 times, converging to near-optimality within 150 episodes (Figure~\ref{average_steps_to_win}). \textcolor{black}{Similar observations were made for Case II. For instance, in Prior 2 with $\Phi_{max} = 0.33$, the agent managed to increase the number of successful episodes from $17.6$ to $444.7$ while slightly increasing, as expected, the number of failures from $14.5$ to $38.9$. }

Finally, we tested Q-learning. 
%Specifically, we ran RCRL but without the restriction that action selection must come from $A_{safe}$.
Q-learning had almost no successful crossings of the bridge in the first 500 episodes, so we ran it for 1500 episodes and found that it only converged to a near-optimal policy about half the time, on average entering unsafe states 990.5 times and successfully crossing the bridge 414.6 times.

Table \ref{tab:average_totals} summarizes the number of successes and failures for each agent. To understand better the rate of convergence to near-optimality, \textcolor{black}{Figures \ref{average_steps_to_win}-\ref{average_steps_to_win2}} display the number of steps taken by the agent to cross the bridge at every successful episode (it displays 400 if the agent never crossed the bridge) averaged over the 10 experiments. On each graph we display for comparison the theoretical least number of steps it could cross the bridge in, which is 22. Note that because the BridgeCross is slippery, even an optimal policy would have fluctuations above the 22-steps line.

\newcommand{\y}{0.45}
\begin{figure}[!ht] 
\centering
  \begin{subfigure}[b]{\y\linewidth}
    \centering
    \includegraphics[width=\linewidth]{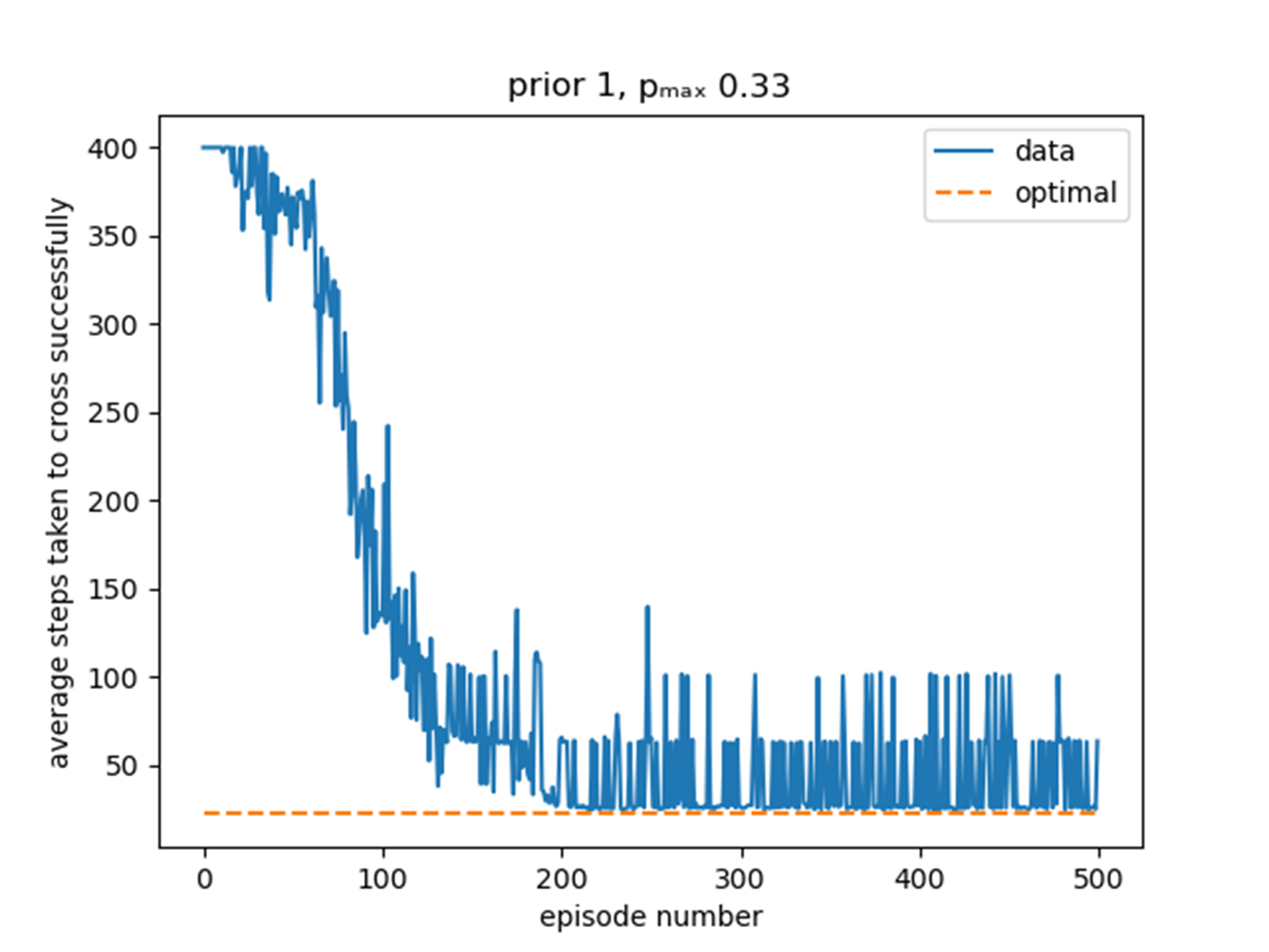} 
    \caption{} 
    \label{fig7:ap} 
    \vspace{3ex}
  \end{subfigure}%% 
  \begin{subfigure}[b]{\y\linewidth}
    \centering
    \includegraphics[width=\linewidth]{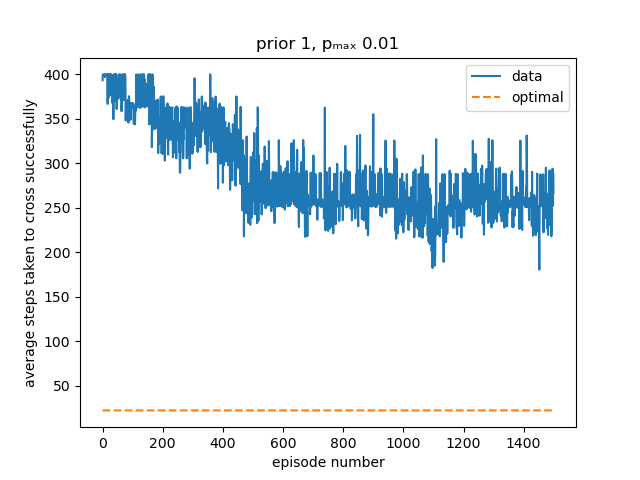}
    \caption{} 
    \label{fig7:bp} 
    \vspace{3ex}
  \end{subfigure} 
  \begin{subfigure}[b]{\y\linewidth}
    \centering
    \includegraphics[width=\linewidth]{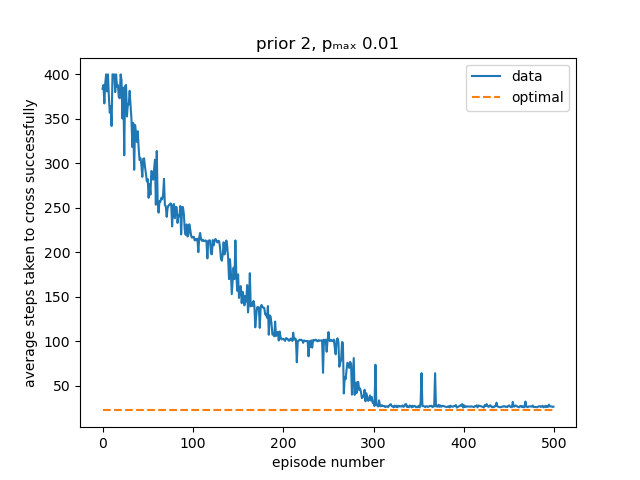} 
    \caption{} 
    \label{fig7:app} 
    \vspace{3ex}
  \end{subfigure}%% 
  \begin{subfigure}[b]{\y\linewidth}
    \centering
    \includegraphics[width=\linewidth]{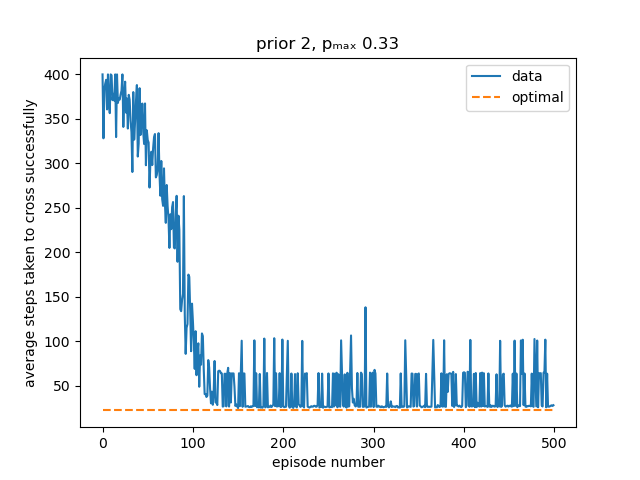} 
    \caption{} 
    \label{fig7:bpp} 
    \vspace{3ex}
  \end{subfigure} 
  \begin{subfigure}[b]{\y\linewidth}
    \centering
    \includegraphics[width=\linewidth]{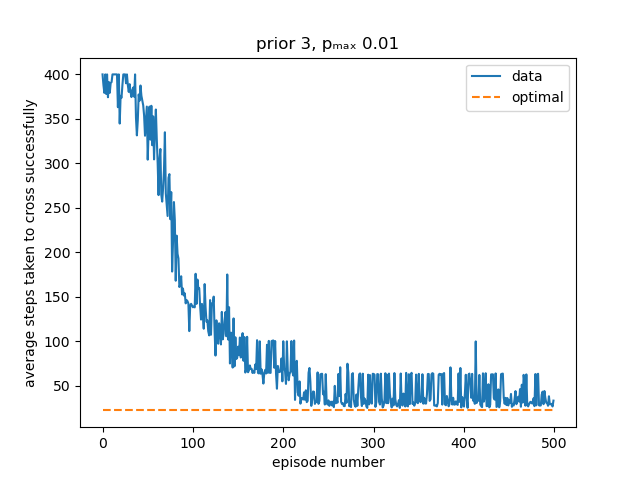} 
    \caption{} 
    \label{fig7:cp} 
  \end{subfigure}%%
  \begin{subfigure}[b]{\y\linewidth}
    \centering
    \includegraphics[width=\linewidth]{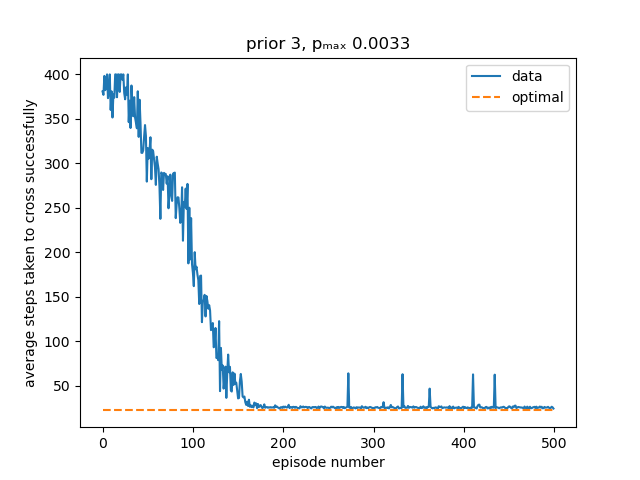} 
    \caption{} 
    \label{fig7:dp} 
  \end{subfigure} 
  \begin{subfigure}[b]{\y\linewidth}
    \centering
    \includegraphics[width=\linewidth]{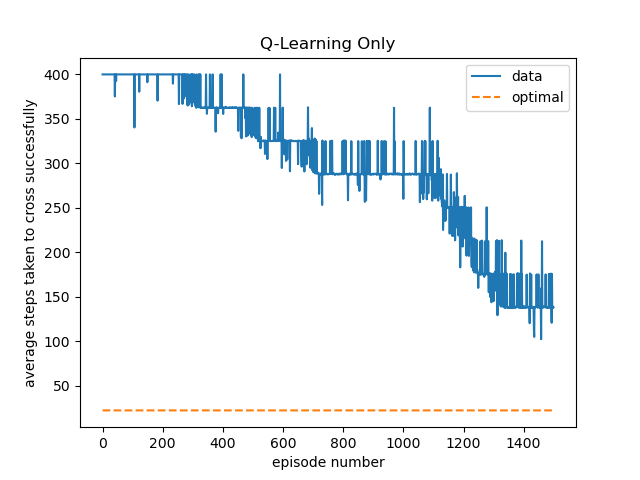} 
    \caption{} 
    \label{fig7:dpp} 
  \end{subfigure} 
  \caption{The number of steps it takes the agent to cross the bridge for every episode where it crosses. Averaged over 10 experiments. Results for Q-learning only and for RCRL across different priors and values of risk  $\Phi_{max}$. As Q-learning converges, it approaches the lower bound on the optimal number of steps per episode.}
  \label{average_steps_to_win} 
\end{figure}

\textbf{Discussions - } The first result of note is how poorly Prior~1 performs with $\Phi_{max} = 0.01$ \textcolor{black}{for both Case I and Case II}. It mostly fails to converge to near-optimal behaviour even with 1500 steps as can be seen \textcolor{black}{in Figure~\ref{average_steps_to_win}b and Figure~\ref{average_steps_to_win2}b,} in fact seeming to converge slower than Q-learning. This occurs because the maximum allowable risk is set too low for the given prior. In particular, there are two main issues with this. The first issue is a type of degenerate behaviour specific to our algorithm and to the completely uninformative prior with overly strict $\Phi_{max}$: given that the agent starts with no information on the transition probabilities, it is unable to tell which actions are safe and which are unsafe. In particular, with $\Phi_{max}$ at $1\%$, the first time the agent arrives at any state $s$ from which it can observe some unsafe state, it immediately goes into safety mode as it judges that the risk of every action is above $1\%$. Since it has no information on which action is safest, it randomly selects an action (assuming the Q-values were initialized to 0). If that randomly-selected action does not take the agent closer to a risky state, then after updating the agent's beliefs about the transition probabilities for that action, it will believe that action is the safest one from that state. Thus every time it encounters that state again, it will \textit{always} select that action, never attempting any other actions. This behaviour can be seen in Figure \ref{state_count_uninformative}. The state $(13,1)$ has been visited significantly more often than any other state. This has occurred because the first time the agent encountered that state, it chose action $stay$, and as above, from then on always chose $stay$ in state $(13,1)$. This would cause the agent to remain in $(13,1)$ until it slipped off of that state. 

The second issue with having such a strict $\Phi_{max}$ could involve any prior. In this case $\Phi_{max}$ is set so low that actions that may be optimal are simply never tested, as the agent's initial belief about those actions causes the expected risk associated with them to always be greater than $\Phi_{max}$. This should not be viewed as an undesirable consequence of the algorithm, but rather as the algorithm working as intended. With the maximum allowable risk level $\Phi_{max}$ set so low, the agent judges that certain actions are riskier than acceptable and therefore does not take them. 
%We would say that this is simply the agent successfully not taking actions that it believes are riskier than acceptable. 
However, this does raise a more general question about the nature of safe learning in general: ensuring safety while learning necessarily means avoiding actions we believe are too dangerous, so if we want any guarantees on safety, then we must accept that the agent may be unable to explore the entire state space. 
 
The second result of note is that Prior 3 performs much less safely than Prior~2 does at $\Phi_{max} = 0.01$. This seems counter intuitive at first, given that Prior~3 is more accurate and more confident than Prior 2. However, the explanation is quite simple. Prior 3 (initially) causes the agent's expected belief to correctly predict that there is only a $1\%$ chance of moving to an unsafe state on a particular step if the agent selects the action to move away from it. On the other hand, Prior 2 causes the agent's expected belief to predict there is a $6.25\%$ chance of this happening. Thus, Prior 3 (correctly) evaluates the risk of moving within 1 step of a risky state as much lower than Prior 2 does. It is likely that at some points in the experiments, the agent with Prior 3 chose to move within 1 step of an unsafe state where an agent starting with Prior 2 (with the same experiences) would have rejected that action as too risky. The agent with Prior 3 would then be at risk of slipping into an unsafe state. In Figure \ref{fig7:a} and \ref{fig7:b}, we can see exactly this happening, where Prior 3 regularly visits state $(13,8)$, which is adjacent to the unsafe state $(12,8)$. Prior 2 instead regularly moves one more state to the right before moving up to row $13$, since $(12,9)$ is safe.

Prior 3 with $\Phi_{max} = 0.0033$ shows how we can make use of a highly accurate prior to guarantee even less risk, and in this case the agent almost never enters unsafe states, while converging faster than any other setup to near-optimality.  %Note that never sinking is an impossible goal here due to the chance of slipping multiple times in a row. 

The final result is that the rate of convergence of the Q-learning agent is much slower on this MDP than the other agents (excluding Prior 1 with the inappropriate $\Phi_{max} = 0.01$). As in Figure \ref{average_steps_to_win}, Q-learning took between 300 and 1500 episodes to converge when it did, and occasionally failed to converge, compared to 150-300 episodes for the four other agents to converge in all 10 experiments. This was even the case for the agent with the completely uninformative prior, with $\Phi_{max} = 0.33$. This is a key result: it shows that not only can RCRL keep the agent safe during learning when possible, it may also direct the agent to explore more fruitful areas of the state-space. In this case study in particular, the Q-learning agent entered unsafe states so often initially that it took many episodes before it was able to access the bridge and find the reward at the other side. Conversely, since the safe agents mostly avoided ``sinking'' situations, they were able to explore much more of the state space on each episode.

\begin{figure}[!ht] 
\centering
  \begin{subfigure}[b]{\y\linewidth}
    \centering
    \includegraphics[width=1\linewidth]{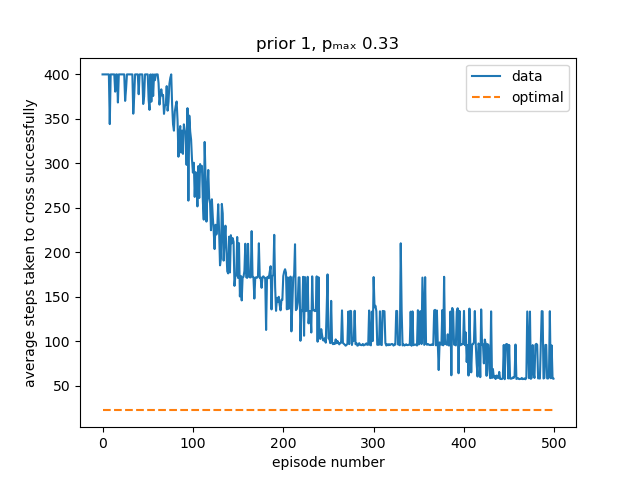} 
    \caption{} 
    \label{fig7_1:ap} 
    \vspace{3.1ex}
  \end{subfigure}%% 
  \begin{subfigure}[b]{\y\linewidth}
    \centering
    \includegraphics[width=1\linewidth]{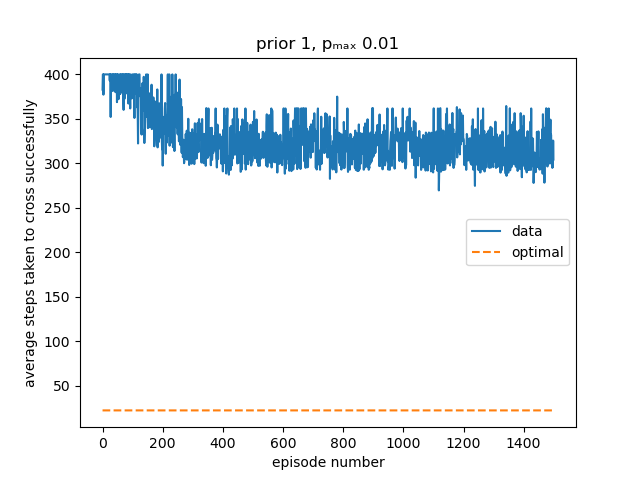} 
    \caption{} 
    \label{fig7_1:bp} 
    \vspace{3.1ex}
  \end{subfigure} 
  \begin{subfigure}[b]{\y\linewidth}
    \centering
    \includegraphics[width=1\linewidth]{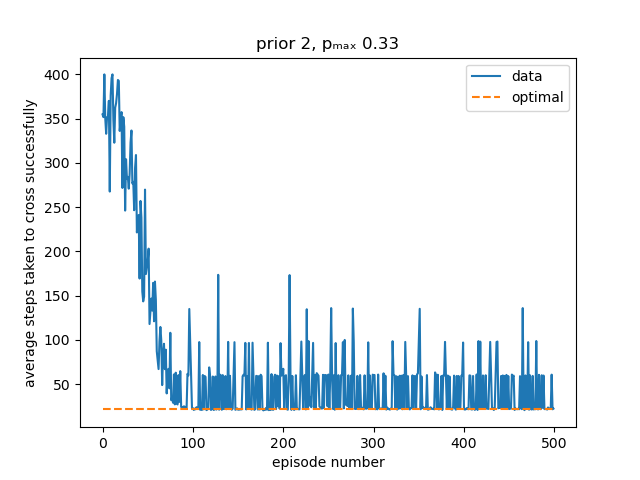} 
    \caption{} 
    \label{fig7_1:app} 
    \vspace{3.1ex}
  \end{subfigure}%% 
  \begin{subfigure}[b]{\y\linewidth}
    \centering
    \includegraphics[width=1\linewidth]{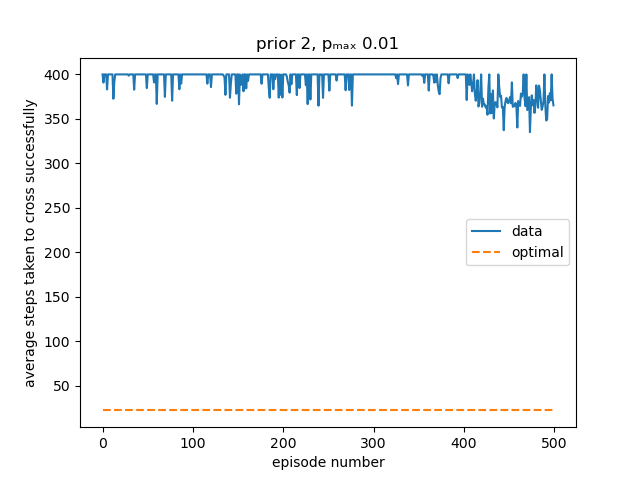} 
    \caption{} 
    \label{fig7_1:bpp} 
    \vspace{3.1ex}
  \end{subfigure} 
  \begin{subfigure}[b]{\y\linewidth}
    \centering
    \includegraphics[width=1\linewidth]{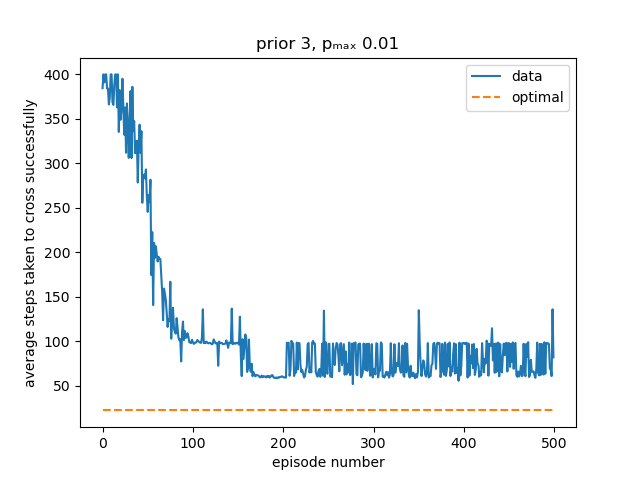} 
    \caption{} 
    \label{fig7_1:cp} 
  \end{subfigure}%%
  \begin{subfigure}[b]{\y\linewidth}
    \centering
    \includegraphics[width=1\linewidth]{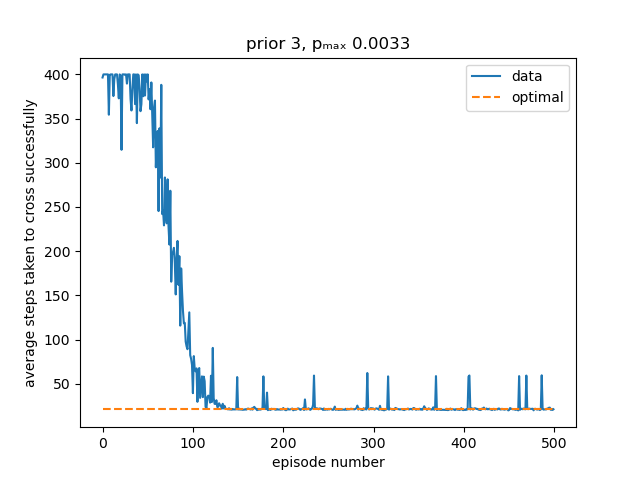} 
    \caption{} 
    \label{fig7_1:dp} 
  \end{subfigure} 
  \begin{subfigure}[b]{\y\linewidth}
    \centering
    \includegraphics[width=1\linewidth]{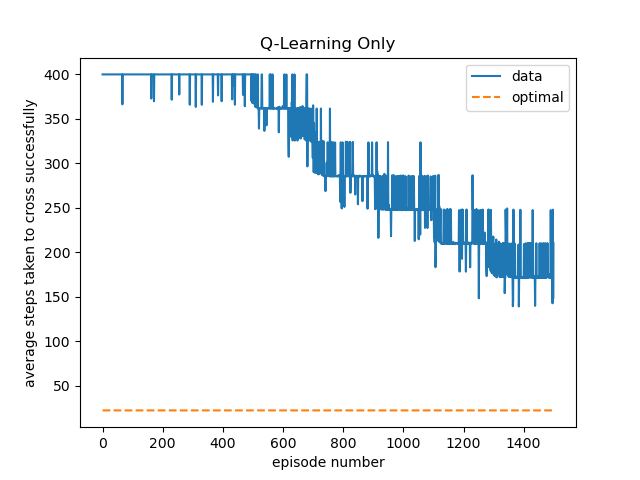} 
    \caption{} 
    \label{fig7_1:dpp} 
  \end{subfigure} 
  \caption{The number of steps it takes the agent to cross the bridge for every episode where it crosses. Averaged over 10 experiments. Results for Q-learning only and for RCRL across different priors and values of risk  $\Phi_{max}$. As Q-learning converges, it approaches the lower bound on the optimal number of steps per episode.}
  \label{average_steps_to_win2} 
\end{figure}

\begin{figure}[!ht]
\centering
\includegraphics[width=0.8\linewidth]{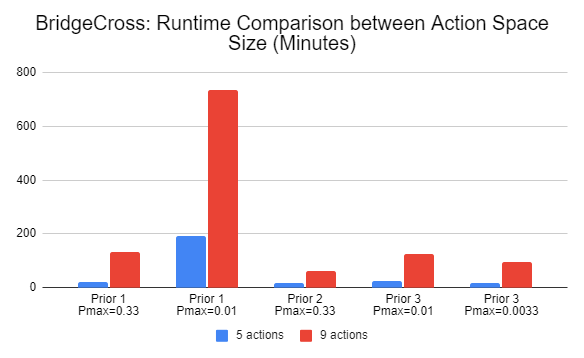}
\caption{Runtime (in minutes) for the training process of BridgeCross case study with different parameters}
\label{fig:runtimes}
\end{figure}

\textbf{Pacman - }
We evaluated the performance of RCRL on a \emph{Pacman} example. Figure \ref{fig:Pacman_setup} depicts the initial state of the environment, where the agent (Pacman) must get to both yellow dots (food) without getting caught by the ghost. Note that because both the agent and the ghost move through the maze, the Pacman MDP has about 10 times more states than the BridgeCross, and up to 5 times more possible next states at any given state. Upon picking up the second piece of food, the agent is given a reward of 1 and the learning episode stops. Every other state incurs a reward of 0 and if the ghost catches Pacman, the learning episode stops with reward 0. The agent has access to five joystick actions at each state, $A = \{\mathit{right}, \mathit{up}, \mathit{left},\mathit{down},\mathit{no\_act}\}$ and will move in the direction selected, or if that direction moves into a wall, then it will stay still. The ghost will with 90\% probability move in the direction that takes it closest to the agent's next location, and with 10\% probability will move in a random direction. For this setup, we assumed an observation \radius $O=3$ and compared two values of the risk horizon, $m=2,3$. We therefore kept constant the other parameters and hyper-parameters: the learning rate $\mu=0.85$; the discount factor $\gamma = 0.9$; the maximum number of steps per episode $max\_steps = 400$; the maximum acceptable risk $\Phi_{max}=0.33$; the prior, which we set to be a completely uninformative prior as in the BridgeCross example; the maximum number of episodes, which we set as 1500 or the number of episodes before the total rate of successful episodes exceeded 75\%.

\newcommand{\xx}{0.45}
\begin{figure}[!t]
\centering
    \begin{subfigure}[b]{\xx\linewidth}
    \centering
    \includegraphics[width=1\linewidth]{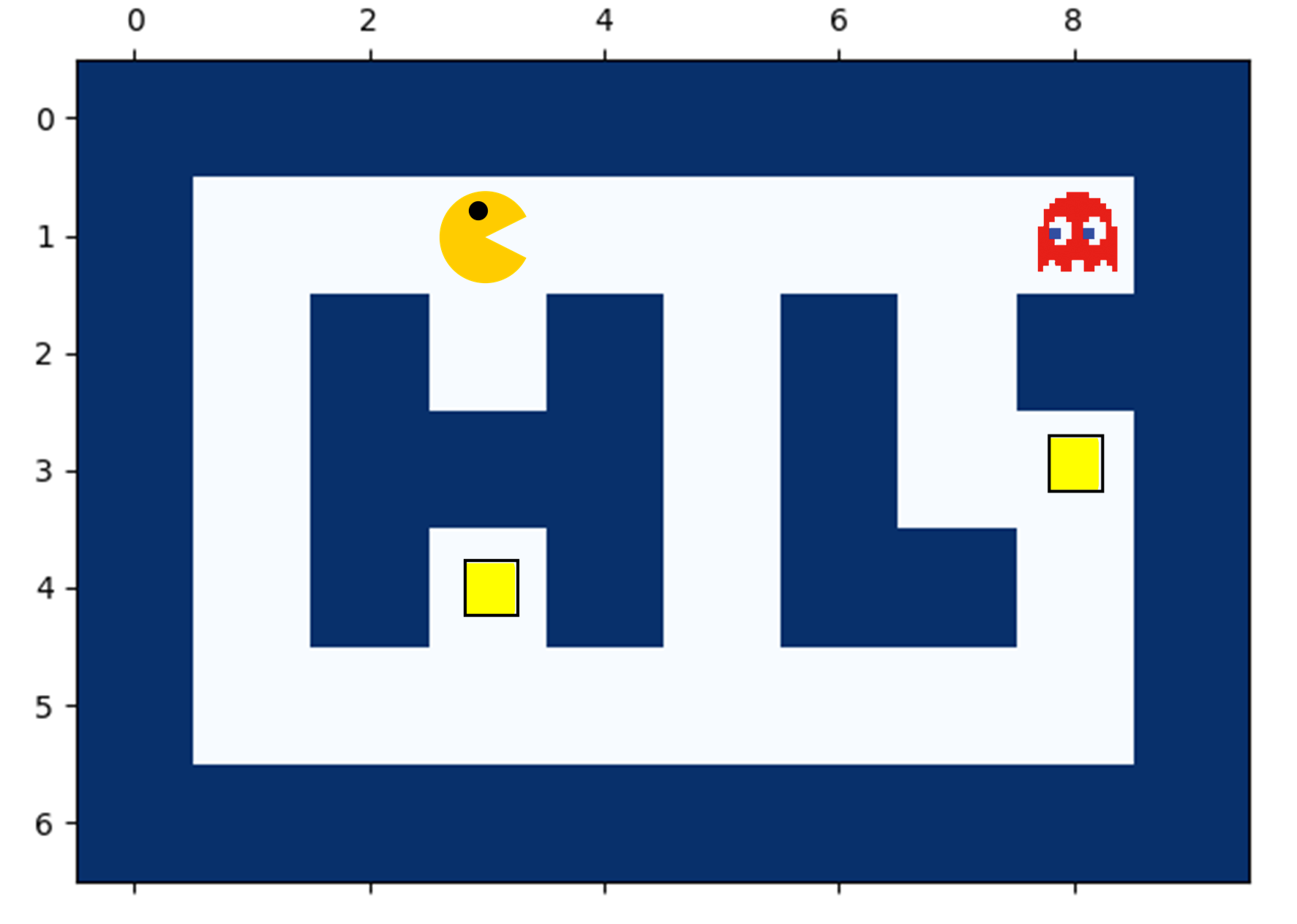} 
    \caption{} 
    \label{fig:Pacman_setup} 
    \vspace{1ex}
  \end{subfigure}%% 
  
  \begin{subfigure}[!t]{\xx\linewidth}
    \centering
    \includegraphics[width=1\linewidth]{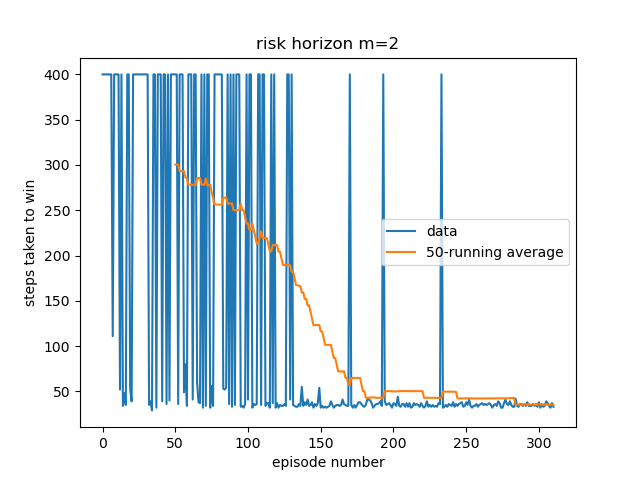} 
    \caption{} 
    \label{fig:Pacman_m2} 
    \vspace{1ex}
  \end{subfigure}%% 
  \begin{subfigure}[!t]{\xx\linewidth}
    \centering
    \includegraphics[width=1\linewidth]{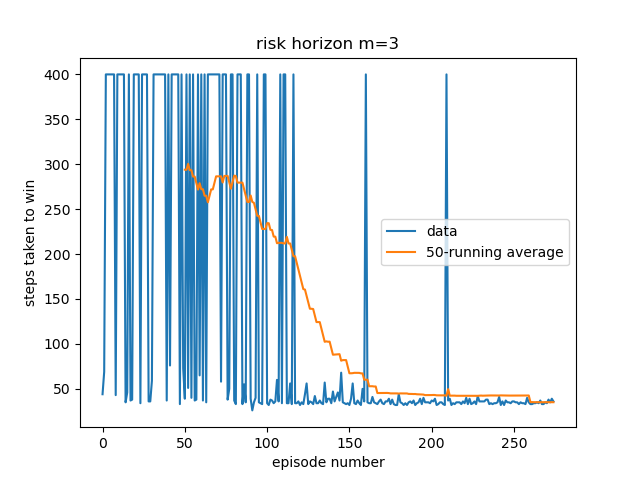} 
    \caption{} 
    \label{fig:Pacman_m3} 
    \vspace{1ex}
  \end{subfigure} 
  \caption{(a) Pacman Setup: agent (Pacman) starts at position (1,3). Food is denoted by yellow dots, and the ghost starts in the top right corner. (b-c) Number of steps taken to win (i.e. eat both foods without being caught by the ghost) on episodes where the agent does win (or 400 if the agent is caught), for risk horizon 2 and 3. The orange line denotes the running average number of steps to win over the previous 50 episodes.}
\end{figure}

As in Table \ref{tab:average_totals}, the agent with a risk horizon of $m = 2$ steps exceeded a success rate of 75\% after 311 episodes, having failed 77 times. The agent with the larger risk horizon of $m = 3$ only took 275 steps to exceed that success rate, and only failed 68 times. Figures \ref{fig:Pacman_m2}-\ref{fig:Pacman_m3} display the number of steps taken by the agent to win (or 400 if they lose) for each agent, as well as the running average number of steps over the previous 50 episodes.

\textbf{Discussion - } The improvement in performance from $m=2$ to $3$ is likely due to the increased foresight of the agent leading it to move away from excessively risky scenarios further in advance, potentially avoiding entering a state from which entering a dangerous state is unavoidable. However, it may also be simply due to the fact that increasing the risk horizon leads to an overall increase in risk estimates, which will naturally cause more actions to be considered too risky and may reduce the number of failures. In other words, we may have been in a situation where decreasing the maximum acceptable risk $\Phi_{max}$ would have led to similar improvements, and the increase in risk horizon was behaving functionally more like a decrease in $\Phi_{max}$. %Research into the potentially differing effects of these two parameters would be interesting.
Both risk-aware agents compare very favourably against the Q-Learning agent, which did not succeed once across 1500 episodes. 

%[NOTE COULD COMMENT ON: Do these converge to near-optimal policies? Use the convergence graphs in Figure \ref{fig:Pacman_convergence} to explain.] This demonstrates, in a second environment and more starkly, how the rate of convergence of Q-learning can be improved by using risk to direct our search of the state-space to potentially more fruitful areas.

%\if\doctype 2
\section{Conclusions}

We proposed a new approach, Risk-aware Cautious Reinforcement Learning (RCRL), to address the problem of safe exploration in MDPs.
A definition of the risk related to taking an action in a given state has made use of the agent's beliefs about the MDP transitions and the safest available actions in future states. 
We have approximated the expectation and variance of the defined risk and have derived a convergence result that justifies the use of those approximations. We have also shown how to derive an approximate bound on the confidence that the risk is below a certain level. 
All these ingredients comprise RCRL, a Safe RL architecture that couples risk estimation and safe action selection with RL. %demonstrating the applicability of this new approach. 
We tested RCRL 
%on a \emph{Slippery BridgeCross} domain, 
and showed that 
%for appropriately chosen values of $\Phi_{max}$, 
it significantly outperform on Q-learning, 
%This was even true for the case in which the agent was given a completely uninformative prior, which meant that it started with no information about the transition probabilities of the MDP. The improvement was 
both in terms of maintaining safety during exploration, as well as of the rate of convergence to an optimal policy. As this approach can be easily interleaved with other RL algorithms we expect similar improvements against other baselines.

% An enticing aspect of this approach is its general applicability. In particular, the combination of the expectation and variance estimation, as well as the use of Cantelli's Inequality to derive an approximate bound on the confidence that the risk is below a certain level, can be decoupled from the rest of the approach and applied more broadly to other RL methods. 
% %We believe there is much scope for exploration in how else these could be used. 
% A direction for future work stems from the idea that optimal policies may be in general unattainable given a certain level of acceptable risk. Rather than treating all risks of a certain level to be the same, we shall work in the direction of attempting to reason about which risks are worth taking in the pursuit of optimality. 

%\fi

\clearpage

\bibliography{biblio}

\onecolumn %% Turn this off if single column is desired for the supplement
% \maketitle
% \vspace{-47mm}
\appendix
\begin{center}
    \LARGE{\textbf{Appendix}}
\end{center}
\section{Convergence Results for the Approximations of the Expected Value and Variance of the Risk}\label{appndix:convergence_results}

\begin{theorem}
Under Q-learning convergence assumptions~\citep{watkins_1989}, namely that reachable state-action pairs are visited infinitely often, the estimate of the mean of the believed risk distribution $\bar \varrho^m(s,a)$ converges to the true risk $\rho^m(s,a)$, and it does so with the variance of the believed risk distribution $\textit{Var}(g^m(s,a)[\mathbf{p}])$ approaching the estimate of that variance $\bar V^m(s,a)$. Specifically, 
\begin{equation*}
    \frac{\left(\bar \varrho^m(s,a) - \rho^m(s,a)\right)}{\sqrt{\bar V^m(s,a)}} \rightarrow \mathcal{N}(0, 1) \text{ in distribution }
\end{equation*}
\end{theorem}

\noindent\textbf{Proof.}\\*
Let us first rewrite the expressions in \eqref{eq:variance_approximation} in vector form, 
first introducing the following covariance matrix for $\mathbf{p}$:
$$
\Sigma = \begin{pmatrix} \textit{Cov}(p^{11}_{b_1}, p^{11}_{b_1}) & \textit{Cov}(p^{11}_{b_1}, p^{12}_{b_1}) &  ... \\
                        \textit{Cov}(p^{12}_{b_1}, p^{11}_{b_1}) & \textit{Cov}(p^{12}_{b_1}, p^{12}_{b_1})\\
   \vdots & &  \ddots \\
   & & & \textit{Cov}(p^{NN}_{b_M}, p^{NN}_{b_M})
\end{pmatrix}. 
$$
Recall that the variables $p^{ij}_a$ are ordered lexicographically by $(i,a,j)$. Here we wrote $b_1$ for the first action in $A$ and $b_M$ for the last one, assuming $|A|=M$. Using matrix $\Sigma$, we can rewrite \eqref{eq:variance_approximation} for the approximate variance as 
\begin{equation}
\textit{Var}(\risk{m}(s,a)) \approx \left(\nabla g^{m}(s,a)[\bar p]\right)^T \Sigma \left(\nabla g^{m}(s,a)[\bar p]\right),~~\nabla g^{m}(s,a)[\bar p] = \left. \begin{bmatrix}  \frac{\partial g^{m}(s,a)}{\partial x^{11}_{b_1}}  \\  \frac{\partial g^{m}(s,a)}{\partial x^{12}_{b_1}} \\ \vdots \\  \frac{\partial g^{m}(s,a)}{\partial x^{NN}_{b_M}}  \end{bmatrix} \right|_{\mathbf{x} = \mathbf{\bar p}},
\end{equation}
where $\nabla g^{m}(s,a)[\bar p]$ is the gradient vector of $g^m(s,a)$ evaluated at $\mathbf{\bar p}$. 

In the following, we employ the `Delta Method' as described in~\citep{casella_berger_2021} to allow us to derive a convergence result for the approximations for the mean and variance of $\varrho^m(s,a)$ that we defined above. Let us introduce a semi-vectorised representation of \eqref{eq:variance_approximation} where we still leverage the fact that covariances across different state-action pairs are 0, i.e.,
$$
\Sigma^i_b = \begin{pmatrix} \textit{Cov}(p^{i1}_b, p^{i1}_b) & \textit{Cov}(p^{i1}_b, p^{i2}_b) &  ... \\
                        \textit{Cov}(p^{i2}_b, p^{i1}_b) & \textit{Cov}(p^{i2}_b, p^{i2}_b)\\
   \vdots & &  \ddots \\
   & & & \textit{Cov}(p^{iN}_b, p^{iN}_b)
                        
\end{pmatrix}
$$
is the variance-covariance matrix for $\left(( p^{ij}_b)_{j=1,...,N}\right)$. Since $\Sigma$ is built by listing the $\Sigma^i_b$ along the diagonal for $i=1,...,N$ and $b \in A$, with zeros elsewhere, we have that \eqref{eq:variance_approximation} can be rewritten as 
\begin{gather}
    \textit{Var}(\risk{m}(s,a)) \approx \sum_{i=1}^N \sum_{b \in A} \left(\nabla^i_b g^{m}(s,a)[\bar p]\right)^T \Sigma \left(\nabla^i_b g^{m}(s,a)[\bar p]\right),~\nabla^i_b g^{m}(s,a)[\bar p] = \left. \begin{bmatrix}  \frac{\partial g^{m}(s,a)}{\partial x^{i1}_b}  \\  \frac{\partial g^{m}(s,a)}{\partial x^{i2}_b} \\ \vdots \\  \frac{\partial g^{m}(s,a)}{\partial x^{iN}_b}  \end{bmatrix} \right|_{\mathbf{x} = \mathbf{\bar p}}~,
\end{gather}
where $\nabla^i_b g^{m}(s,a)[\bar p]$ is the gradient vector $\left(\nabla g^{m}(s,a)[\bar p]\right)$ restricted to entries $\frac{\partial g^{m}(s,a)}{\partial x^{ij}_b}$ for $j = 1,...,N$. We refer to this approximation for the variance of $\varrho^m(s,a)$ as $\bar V^m(s,a)$ ($\approx \textit{Var}(\varrho^m(s,a))$).

Consider the random vector $\mathbf{X} = {(X^{ij}_a)}_{i,j=1,...,N \text{ and } a \in A}$ (with the previously discussed lexicographic order on the $X^{ij}_a$) where each ${(X^{ij}_a)}_{j=1}^N$ follows a Categorical distribution with probabilities $t^{ij}_a$ - i.e. a realisation of the vector $\mathbf{X}$ represents the result of taking one transition from every state-action pair. Wherever $X^{ij}_a = 1$ it represents a transition $s^i \xrightarrow{a} s^j$. $\mathbf{X}$ then has means $\mathbf{t}$ and covariances 
$$
    \textit{Cov}(X^{ij}_a,X^{st}_b) = \begin{cases} -t^{ij}_a t^{st}_b & \text{ if } i=s \text{ and } a=b\\
    0 & \text{ otherwise} 
    \end{cases}
$$
We can then write the variance-covariance matrix for $\mathbf{X}$ as 
$$
\Sigma_{\mathbf{X}\mathbf{X}} = \begin{pmatrix} \textit{Cov}(X^{11}_{b_1}, X^{11}_{b_1}) & \textit{Cov}(X^{11}_{b_1}, X^{12}_{b_1}) &  ... \\
                        \textit{Cov}(X^{12}_{b_1}, X^{11}_{b_1}) & \textit{Cov}(X^{12}_{b_1}, X^{12}_{b_1})\\
   \vdots & &  \ddots \\
   & & & \textit{Cov}(X^{NN}_{b_M}, X^{NN}_{b_M})
                        
\end{pmatrix},
$$

If we observe independent random samples $\mathbf{X}^{(1)}, \mathbf{X}^{(2)},...,\mathbf{X}^{(n)}$ and denote the sample means as $\hat X^{ij}_b = \frac{1}{n} \sum_{k=1}^n (X^{ij}_b)^{(k)}$, or $\mathbf{\hat X} = \frac{1}{n}\sum_{k=1}^n \mathbf{X}^{(k)}$ then for the function $g^n(s,a)\left[\mathbf{x}\right]$ we have, 
%as we did in  \ref{eq:statistical_approximation} for $\mathbf{p}$ and it's mean $\mathbf{\bar p}$,

$$g^m(s,a)[\mathbf{\hat X}] \approx g^m(s,a)[\mathbf{t}] + \sum_{i,j=1}^N \sum_{b \in A} \frac{\partial  g^m(s,a)}{\partial x^{ij}_b} (\hat X^{ij}_b - t^{ij}_b),$$
This is a direct result from the first-order Taylor expansion around $\mathbf{t}$, and therefore the derivatives are evaluated at $\mathbf{t}$. In vector notation, we have 

$$g^m(s,a)[\mathbf{\hat X}] \approx g^m(s,a)[\mathbf{t}] + (\nabla g^m(s,a)[\mathbf{t}])^T(\mathbf{\hat X} - t),$$
where 

\begin{equation*}
    \left(\nabla g^{m}(s,a)[\mathbf{t}]\right) = \left. \begin{bmatrix}  \frac{\partial g^{m}(s,a)}{\partial x^{11}_b}  \\  \frac{\partial g^{m}(s,a)}{\partial x^{12}_b} \\ \vdots \\  \frac{\partial g^{m}(s,a)}{\partial x^{NN}_z}  \end{bmatrix} \right|_{\mathbf{x} = \mathbf{t}}
\end{equation*}
From the `Multivariate Delta Method' theorem \citep{casella_berger_2021}, as long as $$\tau^2 := (\nabla g^m(s,a)[\mathbf{t}])^T \Sigma_{\mathbf{X}\mathbf{X}} (\nabla g^m(s,a)[\mathbf{t}]) > 0,$$ which we will prove later in Lemma~\ref{1} and Lemma~\ref{2}, we have the following convergence:

\begin{equation}
    \sqrt{n} \left( g^m(s,a)[\mathbf{\hat X}] - g^m(s,a)[\mathbf{t}]\right) \rightarrow \mathcal{N}(0, \tau^2) \text{ in distribution}.
\end{equation}
Note that this is equivalent to 
\begin{equation}
\label{eq:delta_method_convergence}
    \frac{\sqrt{n} \left( g^m(s,a)[\mathbf{\hat X}] - g^m(s,a)[\mathbf{t}]\right)}{\tau} \rightarrow \mathcal{N}(0, 1) \text{ in distribution},
\end{equation}
where $\tau := \sqrt{\tau^2}$. %\edit{[why not using $|\tau|$ instead? ROHAN: I felt I had to put this in because I defined tau-squared first, so until we introduce tau, the squared symbol doesn't mean anything. Maybe this is overkill?]}.   

In the following we define $\mathbf{\bar p}^{(n)}$ and $\Sigma^{(n)}$ to be what $\mathbf{\bar p}$ and $\Sigma$ would have been had the agent started with it's prior about the transition probabilities $\mathbf{p}$ and then witnessed exactly the transitions represented by the random sample $\mathbf{X}^{(1)}, \mathbf{X}^{(2)},...,\mathbf{X}^{(n)}$. Formally, suppose that the agent's starting prior was, for each state-action pair $(s^i,b)$, that $p^{i1}_b,p^{i2}_b,...,p^{iN}_b \sim Dir(\alpha^{i1}_b,\alpha^{i2}_b,...,\alpha^{iN}_b)$. Then we can consider the random variables ${p^{i1}_b}^{(n)},{p^{i2}_b}^{(n)},...,{p^{iN}_b}^{(n)} \sim Dir(\alpha^{i1}_b + n\hat X^{i1}_b,\alpha^{i2}_b +n\hat X^{i2}_b,...,\alpha^{iN}_b + n\hat X^{iN}_b)$. Since $n\hat X^{ij}_b$ is the count of the number of times $X^{ij}_b$ was $1$ in the random sample, this new distribution is exactly the result of performing Bayesian inference on the prior given the random sample as our new data. We then let 

\begin{equation*}
    \bar {p^{ij}_b}^{(n)} := \E \left[{p^{ij}_b}^{(n)}\right] = \frac{\alpha^{ij}_b + n\hat X^{ij}_b}{\sum_{k=1}^N \left(\alpha^{ik}_b + n\hat X^{ik}_b\right)},
\end{equation*}
and we also define $\Sigma^{(n)}$ as the covariance matrix of the ${p^{ij}_b}^{(n)}$ over all $i,j,b$, namely
$$
\Sigma^{(n)} = \begin{pmatrix} \textit{Cov}({p^{11}_b}^{(n)}, {p^{11}_b}^{(n)}) & \textit{Cov}({p^{11}_b}^{(n)}, {p^{12}_b}^{(n)}) &  ... \\
                        \textit{Cov}({p^{12}_b}^{(n)}, {p^{11}_b}^{(n)}) & \textit{Cov}({p^{12}_b}^{(n)}, {p^{12}_b}^{(n)})\\
   \vdots & &  \ddots \\
   & & & \textit{Cov}({p^{NN}_z}^{(n)}, {p^{NN}_z}^{(n)})
                        
\end{pmatrix},
$$
From Lemma~\ref{1}, we have

\begin{equation}
\label{eq:proved_later}
    \frac{\sqrt{n} \left( g^m(s,a)[\mathbf{\bar p}^{(n)}] - g^m(s,a)[\mathbf{\hat X}]\right)}{\tau} \rightarrow 0 \text{ in probability},
\end{equation}
and this allows us to use the well-known Slutsky's Theorem~\citep{slutsky} on \eqref{eq:proved_later} and \eqref{eq:delta_method_convergence} to show that 

\begin{equation}
\label{eq:intermediate_convergence_result}
    \frac{\sqrt{n} \left( g^m(s,a)[\mathbf{\bar p}^{(n)}] - g^m(s,a)[\mathbf{t}]\right)}{\tau} \rightarrow \mathcal{N}(0, 1) \text{ in distribution}.
\end{equation}
We must make one more modification to this result. Let   $$(\tau^{(n)})^2 := \left(\nabla g^{m}(s,a)[\mathbf{\bar p}^{(n)}]\right)^T \Sigma^{(n)} \left(\nabla g^{m}(s,a)[\mathbf{\bar p}^{(n)}]\right).$$ 
We would like to show that $n (\tau^{(n)})^2 \rightarrow \tau^2$ in probability. To do this, first note that $\mathbf{\bar p}^{(n)} \rightarrow \mathbf{t}$ in probability, so since $g^m(s,a)$ has continuous derivatives we have that $(\nabla g^m(s,a)[\mathbf{\bar p}^{(n)}]) \rightarrow (\nabla g^m(s,a)[\mathbf{t}])$ in probability. Next we note that  $n \Sigma^{(n)} \rightarrow \Sigma_{\mathbf{X}\mathbf{X}}$ in probability. This is because for the $(i,b_1,j),(s,b_2,t)$-entry we have $0 \rightarrow 0$ if $i \ne s$ or $b_1 \ne b_2$, and otherwise we have 

\begin{align*}
    n \textit{Cov}({p^{ij}_b}^{(n)},{p^{it}_b}^{(n)}) &= \frac{- n(\alpha^{ij}_b +n\hat X^{ij}_b)( \alpha^{it}_b +n\hat X^{it}_b)}{(\sum_{k=1}^N (\alpha^{ik}_b +n\hat X^{ik}_b))^2(1 + \sum_{k=1}^N (\alpha^{ik}_b +n\hat X^{ik}_b))}\\
    &= \frac{- n(\alpha^{ij}_b +n\hat X^{ij}_b)( \alpha^{it}_b +n\hat X^{it}_b)}{(n + \sum_{k=1}^N \alpha^{ik}_b)^2(n + 1 + \sum_{k=1}^N \alpha^{ik}_b)}\\
    &\rightarrow -t^{ij}_b t^{it}_b = \textit{Cov}(X^{ij}_b,X^{it}_b). 
\end{align*}
Therefore we have that the products converge in probability:

\begin{align*}
    n (\tau^{(n)})^2 &=  \left(\nabla g^{m}(s,a)[\mathbf{\bar p}^{(n)}]\right)^T n\Sigma^{(n)} \left(\nabla g^{m}(s,a)[\mathbf{\bar p}^{(n)}]\right) \\
    &\rightarrow (\nabla g^m(s,a)[\mathbf{t}])^T \Sigma_{XX} (\nabla g^m(s,a)[\mathbf{t}]) =  \tau^2.
\end{align*}
Since $\tau^2$ is always positive, and the square root function is therefore continuous at $\tau^2$, we have that $\sqrt{n} \tau^{(n)} \rightarrow \tau$, and so $\frac{\tau}{\sqrt{n} \tau^{(n)}} \rightarrow 1$ in probability. Now we can finally apply Slutsky's Theorem to obtain our final result, which is 

\begin{equation}
    \frac{\left( g^m(s,a)[\mathbf{\bar p}^{(n)}] - g^m(s,a)[\mathbf{t}]\right)}{\tau^{(n)}} \rightarrow \mathcal{N}(0, 1) \text{ in distribution}.
\end{equation}

Recall that $g^m(s,a)[\mathbf{t}]$ is the actual risk in the current state $s$, $g^m(s,a)[\mathbf{\bar p}^{(n)}]$ is the agent's approximation of the expectation of the risk given it's beliefs, and $(\tau^{(n)})^2$ is the agent's approximation of the variance of the risk given it's beliefs (both, in this case, assuming it has seen exactly $n$ transitions from each state). So indeed our estimate of the mean of the believed risk distribution converges to the true risk with enough data, and it does so with the variance of the believed risk distribution approaching our estimate of that variance. 
\medskip
\begin{lemma}\label{1}
Given the definition of the polynomial $g^m(s,a)[\mathbf{x}]$, we have the following:
$$
\frac{\sqrt{n} \left( g^m(s,a)[\mathbf{\bar p}^{(n)}] - g^m(s,a)[\mathbf{\hat X}]\right)}{\tau} \rightarrow 0 \text{ in probability}
$$
\end{lemma}
\noindent\textbf{Proof.}\\*
As required for the convergence results in Theorem \ref{thm:convergence}, one can see that all of the coefficients in $g^m(s,a)[\mathbf{x}]$ are either 0 or 1. This means that we can rewrite it as a sum of terms of the form $$\prod_{i,j,b} \left(x^{ij}_b\right)^{n^{ij}_b}$$ for exponents $n^{ij}_b$. This means that we can write $$\frac{\sqrt{n} \left( g^m(s,a)[\mathbf{\bar p}^{(n)}] - g^m(s,a)[\mathbf{\hat X}]\right)}{\tau}$$ as a sum of terms of the form

\begin{equation*}
    \frac{\sqrt{n}}{\tau}\left(\prod_{i,j,b} \left(\bar {p^{ij}_b}^{(n)}\right)^{n^{ij}_b} - \prod_{i,j,b} \left(\hat X^{ij}_b\right)^{n^{ij}_b} \right).
\end{equation*}
Substituting in the definition of $\bar {p^{ij}_b}^{(n)}$ to this expression yields

\begin{equation*}
    \frac{\sqrt{n}}{\tau}\left(\prod_{i,j,b} \left(\frac{\alpha^{ij}_b + n\hat X^{ij}_b}{\sum_{k=1}^N \left(\alpha^{ik}_b + n\hat X^{ik}_b\right)}\right)^{n^{ij}_b} - \prod_{i,j,b} \left(\hat X^{ij}_b\right)^{n^{ij}_b} \right)
\end{equation*}
And we can simplify this by leveraging that $\sum_{k=1}^N \left(n\hat X^{ik}_b\right) = n$, to get 

\begin{equation*}
    \frac{\sqrt{n}}{\tau}\left(\prod_{i,j,b} \left(\frac{\alpha^{ij}_b + n\hat X^{ij}_b}{n + \sum_{k=1}^N \alpha^{ik}_b}\right)^{n^{ij}_b} - \prod_{i,j,b} \left(\hat X^{ij}_b\right)^{n^{ij}_b} \right)
\end{equation*}

Now, the $\alpha^{ij}_b$ are constants, as is $\tau$, and the values of $\hat X^{ij}_b$ are all bounded between $0$ and $1$. Thus to show that this expression converges to $0$ in probability, we will write it as one quotient, and show that some term in the denominator dominates all terms in the numerator. Let $M := \sum_{i,j,b} n^{ij}_b$. The expression above is equal to 

\begin{align*}
    \frac{\sqrt{n}}{\tau} \left(\frac{\prod_{i,j,b}\left(\alpha^{ij}_b + n\hat X^{ij}_b\right)^{n^{ij}_b} - \prod_{i,j,b} \left(\hat X^{ij}_b \left(n + \sum_{k=1}^N \alpha^{ik}_b\right) \right)^{n^{ij}_b} }{\prod_{i,j,b} \left(n + \sum_{k=1}^N \alpha^{ik}_b\right)^{n^{ij}_b}}  \right) \\
\end{align*}

Now on the numerator of the inner quotient, there are only two terms of order $n^M$. One is an $$n^M \prod_{i,j,b} \left(\hat X^{ij}_b\right)^{n^{ij}_b}$$ that comes from the product on the left, and one is a $$-n^M \prod_{i,j,b} \left(\hat X^{ij}_b\right)^{n^{ij}_b}$$ 
from the product on the right, and these cancel each other out. This means the numerator is entirely of order $n^{M-1}$ or less. On the other hand, the denominator of the inner quotient contains the term $n^M$. Therefore, even after multiplying by the $\frac{\sqrt{n}}{\tau}$ on the outside, which would mean the highest order term on in the numerator could be as high as $n^{M-\frac{1}{2}}$, the $n^M$ in the denominator still dominates and the expression as a whole will converge to $0$ in probability. Since $$\frac{\sqrt{n} \left( g^m(s,a)[\mathbf{\bar p}^{(n)}] - g^m(s,a)[\mathbf{\hat X}]\right)}{\tau}$$ was a sum of expressions of that form, and they all converge to $0$ in probability, we get the result we desired, which is that 

$$\frac{\sqrt{n} \left( g^m(s,a)[\mathbf{\bar p}^{(n)}] - g^m(s,a)[\mathbf{\hat X}]\right)}{\tau} \rightarrow 0 \text{ in probability}$$

\begin{lemma}\label{2}
The defined variable $\tau^2 := (\nabla g^m(s,a)[\mathbf{t}])^T \Sigma_{\mathbf{X}\mathbf{X}} (\nabla g^m(s,a)[\mathbf{t}])$ is strictly greater than zero, namely $\tau^2 > 0$.
\end{lemma}
\noindent\textbf{Proof.}\\*
Note that the covariance matrix can be written as $\Sigma_{\mathbf{X}\mathbf{X}} = \E [(\mathbf{X} - \mathbf{t})(\mathbf{X} - \mathbf{t})^T]$ (recall $\mathbf{t}$ is the mean vector for $\mathbf{X}$). So we have

\begin{align*}
     \tau^2 &= \E [(\nabla g^m(s,a)[\mathbf{t}])^T (\mathbf{X}-\mathbf{t})(\mathbf{X}-\mathbf{t})^T (\nabla g^m(s,a)[\mathbf{t}]) \\
     &=  \E [((\nabla g^m(s,a)[\mathbf{t}])^T (\mathbf{X}-\mathbf{t}))^2 ]
\end{align*}

where we note that $s := (\nabla g^m(s,a)[\mathbf{t}])^T (\mathbf{X}-\mathbf{t})$ is a real-valued random variable, so $s^T = s$. Thus to prove $\tau^2 > 0$ we simply have to show that $s \ne 0$ for some value of $\mathbf{X}$ that occurs with non-zero probability.  

Now,
\begin{align*}
    s &= \sum_{i,j,b} \left. \frac{\partial g^m(s,a)}{\partial x^{ij}_b} \right|_{\mathbf{x} = \mathbf{t}} (X^{ij}_b - t^{ij}_b) \\
    &= \sum_{\text{state-action pairs }(s^i,b)} \left( \sum_{\text{possible next states }s^j}  \left. \frac{\partial g^m(s,a)}{\partial x^{ij}_b} \right|_{\mathbf{x} = \mathbf{t}} (X^{ij}_b - t^{ij}_b) \right)
\end{align*}

So let $s^i_b := \sum_{\text{states }s^j}  \left. \frac{\partial g^m(s,a)}{\partial x^{ij}_b} \right|_{\mathbf{x} = \mathbf{t}} (X^{ij}_b - t^{ij}_b)$, then $s = \sum_{\text{state-action pairs } (s^i,b)} s^i_b$.  

We need to show that there is some possible value of $\mathbf{X}$ such that $s \ne 0$. Now the value of $\mathbf{X}$ is determined by the values of $\mathbf{X}^i_b := (X^{ij}_b)_{j=1}^N$ for each state-action pair $(s^i,b)$. Furthermore, these $\mathbf{X}^i_b$ are independent, and the value of $s^i_b$ depends only on the value of $\mathbf{X}^i_b$. So if there is some state-action pair $(s^i,b)$ such that two possible values of $\mathbf{X}^i_b$ yield two distinct values of $s^i_b$ both with nonzero probability, then we can fix the values of the $X^{hj}_{b'}$ for all $j$ and all $(h,b') \ne (i,b)$ to be some values that occur with non-zero probability, which would fix the value of $s - s^i_b$, and so we could use our two distinct values of $s^i_b$ to find two distinct values of $s$. Both cannot be $0$, so we would be done.  

Now, the value of $\textbf{X}^i_b$ is characterized by picking one $j$ s.t. $X^{ij}_b = 1$, and setting all other $X^{il}_b = 0$ for $l \ne j$. This means that to find two different values of some $s^i_b$, we just need to find states $s^i,s^j,s^l$ and an action $b$ such that the derivatives $\left. \frac{\partial g^m(s,a)}{\partial x^{ij}_b} \right|_{\mathbf{x} = \mathbf{t}}$ and $\left. \frac{\partial g^m(s,a)}{\partial x^{il}_b} \right|_{\mathbf{x} = \mathbf{t}}$ are distinct. Then setting $X^{ij}_b = 1$ would yield a different value of $s^i_b$ from setting $X^{il}_b = 1$. So long as the events $X^{ij}_b = 1$ and $X^{il}_b = 1$ both have nonzero probability, we would be done.  

In order to show that such states $s^i,s^j,s^l$ and such an action $b$ exist, we must introduce vectors $A^n$ that will effectively keep track of each state's contribution towards $g^m(s,a)[\mathbf{t}]$ at the $n$th step of the risk backpropagation. First, define the $N$-by-$N$ matrix 
 $P'_n[\mathbf{x}]$ for $n = 0,1,...,m-2$ such that $$(P'_n[\mathbf{x}])_{ij} = \begin{cases}
      1 & \text{if $i=j$ and $s^i$ is unsafe and observed}\\
      0 & \text{if $i\ne j$ and $s^i$ is unsafe and observed}\\
      x^{ij}_{b_{in}} & \text{otherwise}
    \end{cases} $$

where where $b_{in} := \argmin_b \bar R^{n}_{i}(b)$. Define $P'_{m-1}[\mathbf{x}]$ as

 $$(P'_{m-1}[\mathbf{x}])_{ij} = \begin{cases}
      1 & \text{if $i=j$ and $s^i$ is unsafe and observed}\\
      0 & \text{if $i\ne j$ and $s^i$ is unsafe and observed}\\
      x^{ij}_a & \text{otherwise}
    \end{cases} $$

Then the $P'_n[\mathbf{x}]$ represent the transition probabilities used in the calculation of $g^m(s,a)[\mathbf{x}]$. Specifically, we have that 
\begin{itemize}
    \item $g^n(s^k)[\mathbf{x}]$ is the $k$th entry of the vector $(P'_{n-1}[\mathbf{x}]) ... (P'_0[\mathbf{x}]) g^0$ for $n<m$
    \item $g^m(s^k,a)[\mathbf{x}]$ is the $k$th entry of the vector $(P'_{m-1}[\mathbf{x}]) (P'_{m-2}[\mathbf{x}]) ... (P'_0[\mathbf{x}]) g^0$
    \item So the risk at current state $s$, $g^m(s,a)[\mathbf{t}]$, is the $c$th entry of the vector \newline $(P'_{m-1}[\mathbf{t}]) (P'_{m-2}[\mathbf{t}]) ... (P'_0[\mathbf{t}]) g^0$
\end{itemize}

where $g^0$ is the vector with entries $(g^0)(s^k) := \mathds{1}( s^k \text{ is observed and unsafe})$. We can now define the vectors $A^n$ for $n\le m$ by 
\begin{gather*}
    A^n_i := \begin{cases} \left( (P'_{n-1}[\mathbf{t}]) (P'_{n-2}[\mathbf{t}]) ... (P'_0[\mathbf{t}]) g^0 \right)_i &\text{ if $s^i$ is safely reachable from $s$ in exactly $m-n$ steps } \\
    0 &\text{ otherwise}
    \end{cases}
\end{gather*}

Where in this case a state $s^{s_n}$ is defined to be \textit{safely reachable} from the current state $s^{s_0} = s$ in exactly $n$ steps if 
\begin{itemize}
    \item there are states $s^{s_1},s^{s_2},...,s^{s_{n-1}}$ such that each $t^{{s_p}{s_{p+1}}}_{b_{s_1}}> 0$ for actions $b_{s_0} = a$ and $b_{s_k} := \argmin_b \bar R^{m-k-1}_{s_p}(b)$ determined by the agent's expected safest policy, and 
    \item the states $s^{s_1},s^{s_2},...,s^{s_{n-1}}$ are all safe (note that $s^{s_n}$ can still be unsafe)
\end{itemize}

The purpose of these $A^n$ is just to restrict our attention to the states at step $n$ of the backpropagation that actually influence $g^m(s,a)[\mathbf{t}]$. It is easy to see that 

\begin{equation}
    \left((P'_{m-1}[\mathbf{t}]) ... (P'_{n}[\mathbf{t}]) A^n\right)_c = g^m(s,a)[\mathbf{t}] \text{ for every } n=0,1,...,m
\end{equation}

Now we will be able to argue that if $g^m(s,a)[\mathbf{t}]$ is not equal to 0 or 1, there are states $s^i,s^j,s^l$ and an action $b$ such that $t^{ij}_b$ and $t^{il}_b$ are both non-zero (so there is a positive probability of the events $X^{ij}_b = 1$ and $X^{il}_b = 1$) and such that $\left. \frac{\partial g^m(s,a)}{\partial x^{ij}_b} \right|_{\mathbf{x} = \mathbf{t}} > \left. \frac{\partial g^m(s,a)}{\partial x^{il}_b} \right|_{\mathbf{x} = \mathbf{t}}$.

So assume that $g^m(s,a)[\mathbf{t}]$ is not equal to 0 or 1. Let $n_0$ be the largest index such that $A^{n_0}$ contains an entry $(A^{n_0})_l$ that is equal to 0 and such that $s^l$ is safely reachable from $s$ in exactly $m-n_0$ steps - so $(A^{n_0})_l$ is a 0 that came from $ (P'_{m-1}[\mathbf{t}]) ((P'_{m-2}[\mathbf{t}]) ... (P'_0[\mathbf{t}]) g^0)_l $. 

Since $g^m(s,a)[\mathbf{t}]$ is not 0, $n_0 < m$, and since $s^l$ is safely reachable in $m-n_0$ steps, let $s = s^{s_0}, s^{s_1},...,s^{s_{m-n_0}} = s^l$ be a path along which $s^l$ is safely reachable. Then let $s^i = s^{s_{m-n_0 - 1}}$, and we have that $s^i$ is safe, and $t^{il}_{b_{s_{m-n_0 - 1}}} > 0$. For brevity, write $b' := b_{s_{m-n_0 - 1}}$

Now since $s^i$ is safely reachable in $m-(n_0+1)$ steps, $(A^{n_0 + 1})_i$ cannot be equal to 0 (since $n_0$ was maximal), so there must be some state $s^j$ such that $t^{ij}_{b'} > 0$ and $A^{n_0}_j > 0$, (in order for the term $t^{ij}_{b'} A^{n_0}_j$ to contribute some positive value to $A^{n_0 + 1}_i$). 
Finally, let $p$ be the probability of safely entering $s^i$ in $m-(n_0+1)$ steps (i.e., the sum over all paths that safely reach $s^i$ of the probability of taking that path by choosing the actions specified by the agent's expected safest policy). Then by the chain rule,

\begin{align*}
\left. \frac{\partial g^m(s,a)}{\partial x^{ij}_{b'}} \right|_{\mathbf{x} = \mathbf{t}} = p \left(1 \times A^{n_0}_j + t^{ij}_{b'} \times \left. \frac{\left( (P'_{n_0-1}[\mathbf{x}]) ... (P'_0[\mathbf{x}]) g^0 \right)_j}{\partial x_{ij}}\right|_{\mathbf{x}=\mathbf{t}} \right) > 0
\end{align*}

since clearly $\left. \frac{\left( (P'_{n_0-1}[\mathbf{x}]) ... (P'_0[\mathbf{x}]) g^0 \right)_j}{\partial x_{ij}}\right|_{\mathbf{x}=\mathbf{t}}$ cannot be negative. On the other hand,

\begin{align*}
\left. \frac{\partial g^m(s,a)}{\partial x^{il}_{b'}} \right|_{\mathbf{x} = \mathbf{t}} &= p \left(1 \times (A^{n_0})_l + t^{il}_{b'} \times \left. \frac{\left( (P'_{n_0-1}[\mathbf{x}]) ... (P'_0[\mathbf{x}]) g^0 \right)_l}{\partial x^{il}_{b'}}\right|_{\mathbf{x}=\mathbf{t}} \right) \\
&= p \left(1 \times 0 + t^{il}_{b'} \times \left. \frac{\left( (P'_{n_0-1}[\mathbf{x}]) ... (P'_0[\mathbf{x}]) g^0 \right)_l}{\partial x^{il}_{b'}}\right|_{\mathbf{x}=\mathbf{t}} \right) = 0
\end{align*}

since only one of $t^{il}_{b'}$ and $\left. \frac{\left( (P'_{n_0-1}[\mathbf{x}]) ... (P'_0[\mathbf{x}]) g^0 \right)_l}{\partial x^{il}_{b'}}\right|_{\mathbf{x}=\mathbf{t}}$ can be nonzero - if increasing the value of $t^{il}_{b'}$ could increase the value of  $(A^{n_0})_l = \left( (P'_{n_0-1}[\mathbf{t}]) ... (P'_0[\mathbf{t}]) g^0 \right)_l$ from 0 to greater than 0, then $t^{il}_{b'}$  must have been 0 since $\left( (P'_{n_0-1}[\mathbf{t}]) ... (P'_0[\mathbf{t}]) g^0 \right)_l$ is a sum of products of values from $\mathbf{t}$, all of which are non-negative.  

Hence we have found states $s^i,s^j,s^l$ and an action $b'$ such that the derivatives $\left. \frac{\partial g^m(s,a)}{\partial x^{ij}_{b'}} \right|_{\mathbf{x} = \mathbf{t}}$ and $\left. \frac{\partial g^m(s,a)}{\partial x^{il}_{b'}} \right|_{\mathbf{x} = \mathbf{t}}$ are distinct. Hence the claim.

The only detail left to note is that we assumed that $g^m(s,a)[\mathbf{t}]$ is not either equal to 0 or 1. This assumption is reasonable to make, because if it did not hold, then either our agent would be doomed to enter an unsafe state within $m$ steps, or there is no chance of entering an unsafe state within $m$ steps, according to the agent's expected safest actions. Since what matters to us is how the agent manages risk, situations involving risk 1 or risk 0 are irrelevant.

\clearpage
\section{Confidence Bound on the Risk}\label{appndix:confidence_bound}

To estimate a confidence bound on the risk, we appeal to the Cantelli Inequality, which is a one-sided Chebychev bound~\citep{cantelli1929sui}, and states that for a real-valued random variable $\varrho$ with expectation $\E[\varrho]$ and variance $\textit{Var}[\varrho]$, for $\lambda > 0$ we have 

\begin{equation*}
    \Pr(\varrho \le \E[\varrho] + \lambda ) \ge 1 - \frac{\textit{Var}[\varrho]}{\textit{Var}[\varrho] + \lambda^2}
\end{equation*}

If we let $C := 1 - \frac{\textit{Var}[\varrho]}{\textit{Var}[\varrho] + \lambda^2}$, then rearranging we get that $\lambda = \sqrt{\frac{\textit{Var}[\varrho]C}{1-C}}$. Thus for a variable $\varrho$ that represents some sort of risk, and for some value of $0 < C < 1$, we can say 

\begin{equation*}
    \Pr(\varrho \le P ) \ge C
\end{equation*}

where $\Phi := \E[\varrho] + \sqrt{\frac{\textit{Var}[\varrho]C}{1-C}}$. In words, ``there is at least $C$ chance that the risk is at most $\Phi$.'' Alternatively, ``we are at least $\frac{C}{100} \%$ confident that the risk is at most $\Phi$.''

\clearpage
\section{Risk Estimation}\label{appndix:approximation}

To understand what exactly $\bar \varrho^m(s,a)$ is an approximation of, consider instead calculating this risk using the true transition probabilities $t^{kj}_a$, We would get

\begin{align}
 \rho^{n+1}(s^k,a) & :=
    \begin{cases}
      1 & \text{if $s^k$ is observed and unsafe}\\
      \sum^N_{j=1} t^{kj}_a \rho^{n}(s^j) & \text{otherwise}
    \end{cases} \\
  \rho^{n+1}(s^k) & :=
    \begin{cases}
      1 & \text{if $s^k$ is observed and unsafe}\\
      \rho^{n+1} \left(s^k, \argmin_{a \in A} \bar \varrho^{n+1}(s^k,a) \right)  & \text{otherwise}
    \end{cases}\\
\rho^0(s^k) & := \mathds{1}(s^k \text{ is observed and unsafe})
\end{align} 

Note that we crucially still take the minimum risk action $a$ according to the agent's approximation $\bar \varrho^{n+1}(s^k,a)$. In this case, the term $\rho^m(s,a)$ is the true probability of entering an unsafe state after selecting action $a$ in the agent's current state $s$ and thereafter selecting the actions that \textit{the agent currently believes} will minimize the probability of entering an unsafe state over the horizon $m$. $\bar \varrho^m(s,a)$ is the agent's approximation of $\rho^m(s,a)$.

We will later justify the use of $\bar \varrho^m(s,a)$ as an approximation of  $\rho^m(s,a)$, but for now let us consider why it makes sense to define $m$-step risk as $\rho^m(s,a)$. This because the action $a$ that minimizes believed risk is the action that the agent would choose if it was trying to behave as safely as possible, what I will call going into `safety mode'. Consider the motivating example of a pilot learning to fly a remote control helicopter by incrementally expanding the set of actions they feels safe taking. They start by generating just enough lift to begin flying, then immediately land back down again. They repeat this process a few times until they feel that they have a good understanding of how the helicopter responds to this limited range of inputs. Then they take a risk (by either flying a bit higher, or attempting to move horizontally) and once again immediately land. As they repeat this process of taking small risks and landing to remain safe, they begin to expand their comfort zone. At some point after taking a risk, they will feel comfortable just coming back to a hovering position rather than landing, once they have become confident that they can hover safely. This suggests that a natural process for learning to operate in the face of risks is to repeatedly take small risks followed by going into safety mode until back in a confidently safe state. Thus, when calculating how risky an action is, it makes sense to consider the probability of entering an unsafe state given that after the action the agent will enter safety mode. $\rho^m(s,a)$ does exactly this. 

As mentioned earlier, the other reason for defining the risk $\rho^m(s,a)$ in this way is that it makes it possible for the agent to attempt to calculate the risk without having to reason about the inter-dependency between the calculated risk and the agent's future actions. However, it does more than this. We will see in the next section that it in fact allows the agent to view $\bar \varrho^m(s,a)$ as (an approximation of) the expected value of a random variable for the believed risk, where we can also approximate the \textit{variance} of that random variable, allowing for deeper reasoning about action-selection for Safe RL.

\end{document}